\documentclass[a4paper,fleqn]{cas-dc}
\usepackage[numbers]{natbib}
\usepackage{amssymb}
\usepackage{natbib}
\usepackage{enumitem}
\usepackage{pdflscape}
\usepackage{tabularx}
\usepackage{longtable}
\usepackage{pdflscape}
\usepackage{rotating}
\usepackage{array}
\usepackage{hyperref}
\usepackage{afterpage} 
\usepackage{geometry}
\usepackage{booktabs}
\usepackage{amsmath}
\usepackage[ruled]{algorithm2e}
\usepackage[]{algpseudocode}
\usepackage{mathtools}
\usepackage{algcompatible}
\usepackage{algorithm2e}
\usepackage{lipsum}
\usepackage{rotating} 
\usepackage[T1]{fontenc}
\usepackage{fancyhdr}
\usepackage{supertabular,booktabs}
\usepackage{graphics}
\usepackage{lipsum}

\begin{document}



\font\myfont=cmr16 
\title{ \myfont Classification of Dysarthria based on the Levels of Severity. A Systematic Review}

\author[1]{Afnan Al-Ali}[type=editor,
                        auid=000,bioid=1,
                       ]
                       \ead{aa1805360@qu.edu.qa}

\author[1]{Somaya Al-Maadeed}[type=editor,
                        auid=000,bioid=1,
                      ]
\author[1]{Moutaz Saleh}[type=editor,
                        auid=000,bioid=2,
                       ]                       
\author[2]{Rani Chinnappa Naidu}[type=editor,
                        auid=000,bioid=3,
                      ]
 \author[2]{Zachariah C Alex}[type=editor,
                        auid=000,bioid=3,
                      ]
\author[2]{Prakash Ramachandran}[type=editor,
                        auid=000,bioid=3,
                      ]
\author[3]{Rajeev Khoodeeram}[type=editor,
                        auid=000,bioid=3,
                      ]
\author[2]{Rajesh Kumar M}[type=editor,
                        auid=000,bioid=3,
                      ]                      
    
\credit{Conceptualization of this study, Methodology, Software}

\affiliation[1]{organization={ Computer Science and Engineering Department},
    addressline={Qatar University}, 
    city={Doha, Qatar}
    }
\affiliation[2]{organization={ Vellore Institute of Technology},
    addressline={ Vellore}, 
    city={India}}
 \affiliation[3]{organization={ Université Des Mascareignes},
    addressline={ Mauritius}}

\cortext[aa1805360@qu.edu.qa]{Corresponding author}

\begin{abstract}
Dysarthria is a neurological speech disorder that can significantly impact affected individuals' communication abilities and overall quality of life. The accurate and objective classification of dysarthria and the determination of its severity are crucial for effective therapeutic intervention. While traditional assessments by speech-language pathologists (SLPs) are common, they are often subjective, time-consuming, and can vary between practitioners. Emerging machine learning-based models have shown the potential to provide a more objective dysarthria assessment, enhancing diagnostic accuracy and reliability. This systematic review aims to comprehensively analyze current methodologies for classifying dysarthria based on severity levels. Specifically, this review will focus on determining the most effective set and type of features that can be used for automatic patient classification and evaluating the best AI techniques for this purpose. We will systematically review the literature on the automatic classification of dysarthria severity levels. Sources of information will include electronic databases and grey literature. Selection criteria will be established based on relevance to the research questions. Data extraction will include methodologies used, the type of features extracted for classification, and AI techniques employed. The findings of this systematic review will contribute to the current understanding of dysarthria classification, inform future research, and support the development of improved diagnostic tools. The implications of these findings could be significant in advancing patient care and improving therapeutic outcomes for individuals affected by dysarthria.
\end{abstract}
\maketitle{}
\section{Introduction}
Speech is a distinctive, intricate, dynamic motor activity that enables us to articulate our thoughts and emotions and interact with and regulate our surroundings.
Furthermore, it constitutes one of the five heritable verbal traits, encompassing speech, language, reading, writing, and spelling. Speech involves integrating neurocognitive processes for organizing thoughts into language, motor speech planning for executing verbal messages, and neuromuscular execution for coordinating speech muscles. Together, these processes constitute motor speech activities.

When neurologic impairments affect these motor speech activities, a speech disorder will result, which can also be known as Motor speech disorder (MSD) \cite{shriberg2019estimates}\cite{duffy2019motor}.

There are two main types of MSD: dysarthrias and apraxia of speech. Dysarthria is a group of neurologic speech disorders involving abnormalities in speech production's movement aspects. These disorders manifest as changes in strength, speed, range, steadiness, tone, or accuracy of movements required for breathing, phonation, resonance, articulation, or prosody. Sensorimotor abnormalities, such as weakness, spasticity, incoordination, involuntary movements, or variations in muscle tone, underlie dysarthria. The condition is specifically neurologic and can be categorized into distinct types based on perceptual characteristics and underlying neuropathophysiology. It is essential to define dysarthria accurately to differentiate it from other speech and language disorders and to ensure its meaningful application in research and clinical settings \cite{duffy2019motor}. While Verbal apraxia, which is also known as apraxia of speech, is a contentious condition that some view as a deficit in the motor planning of speech. This disorder is marked by "higher order" errors, including metathesis and segment addition, alongside errors that suggest a lack of coordination in articulation. These characteristics suggest a relatively significant level of damage to the neural system \cite{kent1983acoustic}.

 Traditionally, dysarthria is classified into several subtypes, including spastic, flaccid, hypokinetic, hyperkinetic, ataxic, and mixed. More recently, additional subtypes such as unilateral upper motor neuron dysarthria and undetermined dysarthria have been recognized \cite{laganaro2021sensitivity}. 
Dysarthria can occur at any stage of life. Common causes encompass stroke, severe head injury, brain tumors, Parkinson's disease, multiple sclerosis, motor neuron disease, cerebral palsy, Down's syndrome, and certain medications, including those used to treat epilepsy, which may induce a side effect \cite{duffy2019motor}.
The severity of dysarthria is determined by the degree of involvement in the affected body regions caused by the underlying condition. Assessment of dysarthria means mainly grading its severity. This is typically performed by speech-language pathologists (SLPs) using some descriptive terms, but it can be a time-consuming and labor-intensive process with variations between different SLPs. Thus, it is essential to have objective methods of evaluating the level of intelligibility in dysarthria cases \cite{chandrashekar2019breathiness}. For this purpose, machine learning-based models were developed for automatic assessment of dysarthrias' levels of severity to achieve enhanced diagnostic accuracy, consistency, and reliability, all while maintaining cost-effectiveness and expediency \cite{tong2020automatic}.
In both paths, features are extracted from the candidate samples to help the underlying system for classification, where there are specific sets of features for each type of dysarthria.
To our knowledge, this is the first comprehensive review that analyzes the works of classifying dysarthria cases based on the levels of severity. In the literature, several gaps warrant further research in conducting a systematic review on classifying dysarthria based on severity levels. These gaps include the absence of a review focused explicitly on severity-based classification, limited research on severity levels within specific populations \cite{noffs2018speech}\cite{gandhi2020scoping}\cite{mackenzie2011dysarthria}\cite{lee2021assessment} (population-specific considerations), a lack of standardized measures for assessing severity \cite{gandhi2020scoping}\cite{finch2020speech}\cite{whillans2022systematic}\cite{park2020effect} (measurement challenges), the integration of technology in severity classification \cite{shanmugam2021critical} (technological advancements), a need for a comparative analysis of existing classification systems \cite{rowe2022characterizing} (evaluation of existing approaches), specific causes of dysarthria \cite{chiaramonte2021systematic} (etiological factors), general treatment helping clinicians for stable dysarthria \cite{palmer2007methods} (treatment approaches) and automated intelligibility assessment \cite{huang2021review} (technology-driven assessment methods). Addressing these gaps through a systematic review would offer valuable insights for clinicians, researchers, and stakeholders involved in dysarthria assessment and treatment.

In this systematic review, we aim to answer the below research questions:

\begin{enumerate}
    \item 
        {What are the optimal set and type of features that are consistently effective across various severity levels of dysarthria, enabling automated classification of patients?}
    \item 
    {Which artificial intelligence techniques are most suitable for accurately classifying dysarthria patients, considering factors such as training time efficiency and high accuracy?}
    
\end{enumerate}

The rest of the paper is organized as follows: Section 2 will show the strategy followed in our research, Section 3 presents how dysarthric cases are classified based on clinical techniques, Section 4 will show the same for Section 3, but based on machine learning techniques, Section 5 discusses the results shown in the two previous sections and analyses them. Section 6 refers to some common limitations of the related work. Section 7 suggests a few points to solve the gap in this research area, and finally, the Conclusion in Section 8 will summarize our work and highlights our findings in this review.

\section{Search Strategy}
A comprehensive exploration was conducted on various electronic databases, including ACM, EMBASE, SpringerLink, PubMed, Scopus, IEEE, MDPI, Elsevier, and some other conferences popular in the area, based on the search keywords: "classification" AND "Dysarthria" AND "severity levels" or "Assessment" AND "Dysarthria" AND "Intelligibility." 
Figure 1 shows the search strategy process.

A total of $978$ publications were found. After deleting the duplicates, $733$  publications passed through a screening process where only the articles which contained the keywords in their titles, abstracts, and keywords were selected, and the excluded publications were based on the following criteria:(1) reviews and surveys are not considered, (2) articles do not have the main keywords in their titles and abstracts, (3) general chapters related to motor speech disorders are not considered, and (4) other types of disorders like Dysphagia or Dementia. 

The second stage of the study search is knowing the eligibility of each publication to be selected for the analysis later on in this study. The exclusion criteria in this stage are (1) The tools or software for rehabilitation or therapy, (2) automatic recognition systems or the identification of dysarthria, (3) reviews and surveys related to dysarthria or one of its specific types or causes, (4) the impact of specific features of dysarthric cases, and (5) the binary classification of dysarthria.

The final publications are $44$ articles mainly related to classifying dysarthria patients based on severity levels and assessing the intelligibility factor as severity levels.

\begin{figure}
  \centering
  \includegraphics[width=.5\textwidth]{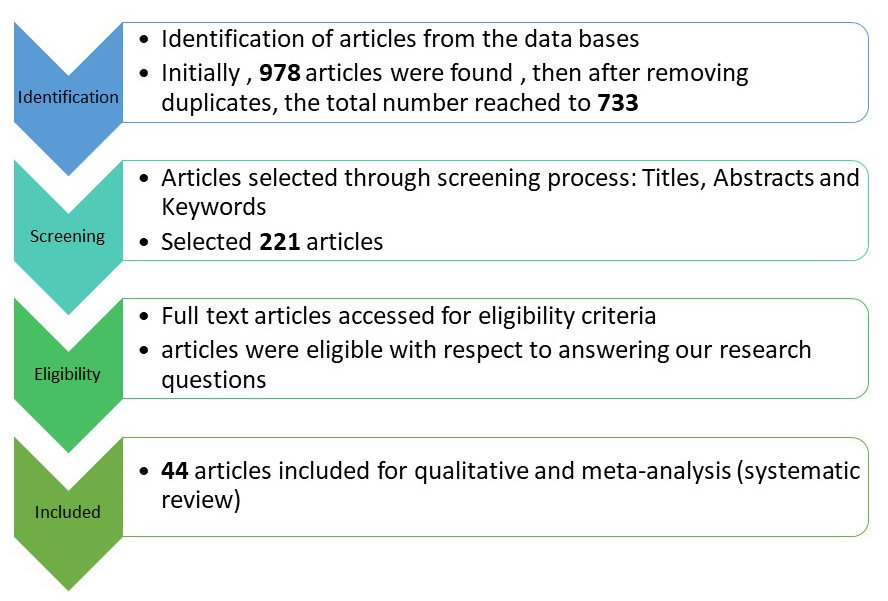}
  \caption{Search Study Process.}
  \label{fig:figure_label2}
\end{figure}

As our primary goal is to classify dysarthric cases based on severity levels, we will refer to both types: clinical or human-based and AI or machine learning-based techniques. We will highlight the methods, extracted features, and the datasets they evaluated the models on.
Figure 2 shows the taxonomy of our review in classifying dysarthria based on the severity levels.

\section{Classification based on Clinical Techniques}
When a dysarthric patient is admitted to the hospital, a specific procedure will be followed based on the clinical setting, available resources, and individual needs. The main points regarding each procedure are explained in the sub-sections below:
\subsection{Methods}
As a clinical procedure, speech-language pathologists (SLPs) follow up on dysarthrias cases. Other rehabilitation professionals, physicians, and nurses may also be involved in such treatment. Each SLPs will search for some descriptive patterns in the patient to confirm his diagnosis and decide the type of treatment based on the severity of the case \cite{stipancic2021you}.

The procedure followed by speech-language pathologists (SLPs) to measure the severity of dysarthria in a patient typically involves employing a comprehensive approach to assess dysarthria in patients. This process includes gathering the patient's medical history to provide context, visually observing their speech and oral motor movements, making perceptual assessments of speech characteristics and severity, evaluating speech intelligibility through tests and measurements, assessing the impact of dysarthria on functional communication abilities, and collaborating with patients, family members, and healthcare professionals to gain a comprehensive understanding of the condition. By following this systematic approach, SLPs can develop tailored treatment plans to address the specific challenges faced by each patient\cite{duffy2019motor} \cite{mcneil2017apraxia}.
\subsection{Features}
Speech-language pathologists (SLPs) use various features to assess the severity of dysarthria in patients. These features include articulation, by assessing the accuracy and precision of speech sound production; phonation, by evaluating voice quality, pitch, loudness, and presence of abnormalities; resonance, by examining the control of the velopharyngeal mechanism for speech clarity; prosody, by analyzing rhythm, stress, intonation, and melodic contour; speech rate, by observing speed, pacing, and pauses; and intelligibility, by determining the percentage of intelligible speech. 

SLPs employ clinical observation, perceptual judgments, and instrumental assessments like acoustic analysis to assess these features. By considering these aspects, SLPs can determine dysarthria severity and develop customized treatment plans to address patients' specific speech challenges \cite{stipancic2021you}\cite{barkmeier2017speech}.

\subsection{Evaluation}
Several standardized rating scales and perceptual judgments are commonly used to assess dysarthria's severity and specific features. These include formal and informal assessments. The formal ones are represented by the most famous Frenchay Dysarthria Assessment (FDA) \cite{enderby1980frenchay}, which evaluates respiration, phonation, articulation, and prosody; other measurements like the  Assessment of Intelligibility of Dysarthric Speech (AIDS), which measures overall speech intelligibility \cite{yorkston1984assessment} and dysarthria profile \cite{robertson1987dysarthria}; the 
 and Voice Handicap Index \cite{jacobson1997voice}. Informal assessments, such as oral motor examinations \cite{freed2018motor}, are often used with formal assessments. Perceptual assessment \cite{blaustein1983reliability} is used by speech-language pathologists, relying on their expertise in active listening and analyzing speech. It's important to note that these perceptual judgments are subjective, emphasizing the need for skilled clinicians to accurately assess and interpret the speech characteristics of individuals with dysarthria. Finally, Communication Activities of Daily Living-Second Edition (CADL-2) \cite{al2014validation}, Although not specific to dysarthria, this assessment measures functional communication abilities in daily life situations. It evaluates the impact of dysarthria on communication in various contexts\cite {altaher2019report}.

 \begin{figure*}
  \centering
  \includegraphics[width=0.9\textwidth]{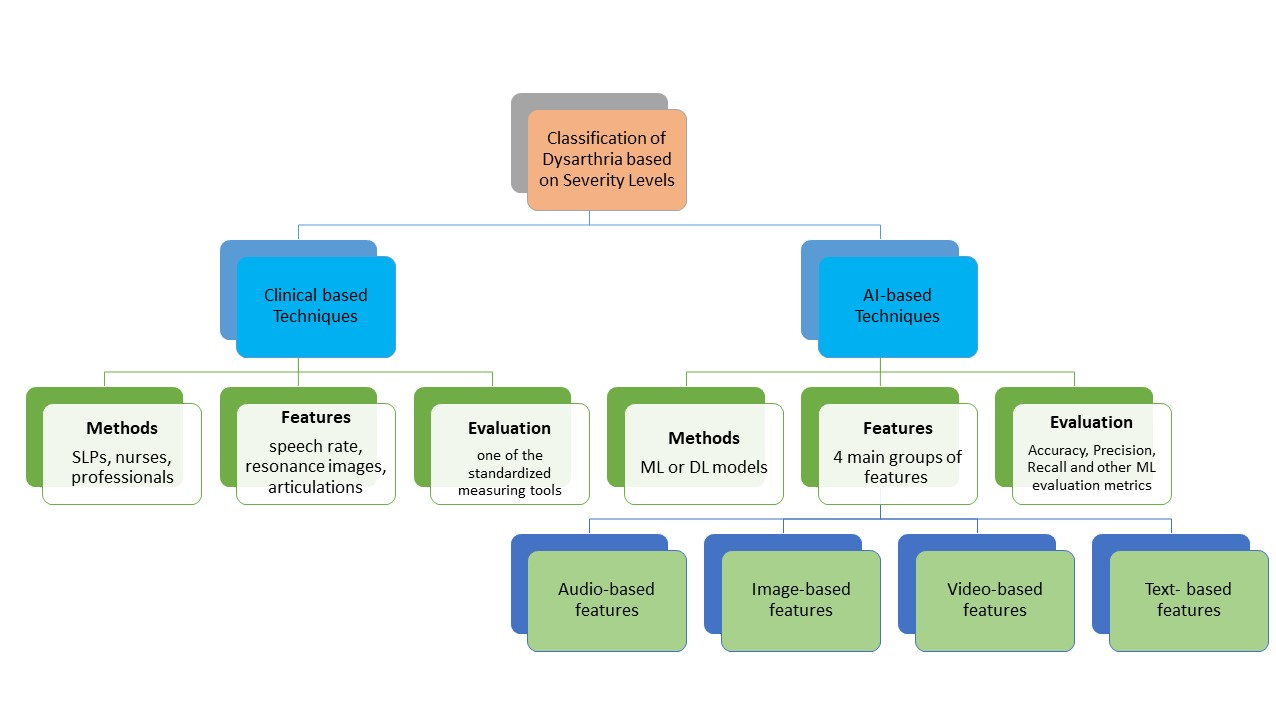}
  \caption{Taxonomy of the study.}
  \label{fig:figure_label}
\end{figure*}

The inclusion of the section on clinical assessment in our review serves two essential purposes. Firstly, it provides a comprehensive understanding of the traditional approaches and methods used in dysarthria classification based on severity levels. By examining the techniques employed by clinicians and the features they consider, we gain insights into the established practices and evaluation tools that have been relied upon for decades. This knowledge is crucial for contextualizing and appreciating the advancements brought by AI-based techniques.

Secondly, incorporating the clinical assessment perspective allows for a comparative analysis between human-based approaches and AI-based methods. By juxtaposing the strengths and limitations of clinical judgment with the capabilities of AI models, we can critically evaluate the potential of AI to augment and enhance dysarthria classification. This comparative analysis enables us to identify the unique contributions of AI techniques, such as increased objectivity, scalability, and potential for automated assessment.
\section{Classification based on AI Techniques}
Despite the effectiveness of SLPs' efforts, this process is time-consuming, mainly subjective, and suffers from differences among individual opinions. This gap inspired the researchers to seek more accurate models, away from the subjective perspective, using AI-based models, as explained below.
\subsection{Methods}
Machine learning and deep learning-based models have shown promise in assessing the severity of dysarthria, offering alternative approaches to traditional assessments conducted by SLPs. These models leverage the power of artificial intelligence to analyze speech patterns and extract meaningful information for severity evaluation \cite{godino2004automatic}.

Several common machine learning approaches, such as support vector machines (SVM), Random Forests (RF), or Artificial Neural Networks (ANN), have been utilized to develop predictive models \cite{hernandez2020prosody}. These models are trained on a dataset of acoustic features or other forms of acoustic features extracted from speech samples of individuals with dysarthria and corresponding severity ratings provided by SLPs. The models learn patterns and relationships between these features and the severity ratings, enabling them to predict/classify the severity of dysarthria in new, unseen cases.

Deep learning, a subset of machine learning, has gained attention in dysarthria severity assessment. Deep Neural Networks (DNN), specifically Long-Short Term Memory (LSTM)\cite{gallardo2021auditory} and convolutional neural networks (CNNs) \cite{vasquez2018multitask}, have been employed to analyze speech signals and capture intricate temporal and spectral patterns. These models can process raw speech data directly or extract features automatically through multiple layers of computation. They can learn complex representations and make predictions based on the learned patterns, allowing for more accurate severity assessment.

Research studies have demonstrated the potential of these machine learning and deep learning models' potential in objectively quantifying dysarthria's severity. These models offer advantages such as increased efficiency, consistency, and potential for remote or self-administered assessments. However, it is essential to note that these models are still evolving. Their performance may vary depending on factors such as the quality and diversity of training data, feature selection, and the complexity of the dysarthria cases \cite{parisi2018feature}\cite{xu2023dysarthria}.
\subsection{Features}
Machine learning and deep learning models for assessing the severity of dysarthria utilize a range of features extracted from speech signals. These features capture different aspects of speech production and can provide valuable information for severity assessment. We can categorize these features into four main groups: audio-based, image-based, video-based, and text-based. Let's break it down:
\subsubsection{Audio-based features}
Audio-based features for dysarthria classification based on severity can be broadly categorized into acoustic, prosodic, and spectral features \cite{kent1999acoustic, sapir2010formant,kim2011vowel, orozco2014new}.

\textit{Acoustic features} include the fundamental frequency (F0), which corresponds to the pitch of the speech, as well as measures of variability such as jitter and shimmer, reflecting the stability of vocal fold vibration. The Harmonics-to-Noise Ratio (HNR) is another acoustic feature that indicates the ratio of energy in the harmonics of the speech signal to the energy in the noise. Lower HNR values may indicate poor voice quality.

\textit{Prosodic features} play a role in dysarthria classification. Speech rate, which refers to the number of syllables per unit of time, can be affected by the severity of dysarthria. Pause duration, or the length of pauses between speech segments, can provide insights into difficulties in speech production or planning. Additionally, changes in stress patterns, including variations in intensity, duration, and pitch of syllables, can indicate the presence of dysarthria.

\textit{Spectral features} offer valuable information for dysar-thria classification. Mel-Frequency Cepstral Coefficients
 (MFCCs) capture the speech signal's short-term power spectrum, providing insights into the speaker's vocal tract shape and function. Linear Predictive Coding (LPC) coefficients represent the spectral envelope of the speech signal, offering information about the vocal tract shape and articulatory movements. Formant frequencies, such as F1, F2, F3, etc., represent the resonant frequencies of the vocal tract during speech production, revealing details about the shape and position of the articulators (tongue, lips, etc.).

The audio-based features mentioned above are not the only types in each group. They represent typical features for dysarthria classification, but additional features can be within each category \cite{freed2018motor}\cite{tartarisco2021artificial}. Table 1 summarizes some studies that used this type of feature for classifying dysarthria based on severity levels or estimating intelligibility for severity classification of dysarthria. 

\subsubsection{Image-based features}
Image-based features in the context of dysarthria analysis involve spectrograms, which are visual representations of the frequency content of an audio signal over time. 

Spectrograms provide a 2D visual representation of the frequency content of an audio signal over time. They are commonly used in speech and audio analysis, including dysarthria classification. By analyzing spectrograms, various features can be extracted to characterize dysarthric speech. These features include spectral envelope, spectral patterns, spectral variations, and spectral entropy. Spectrograms offer valuable insights into the frequency components and dynamics of speech signals \cite{kempler2002effect}.

Mel spectrograms, also known as Mel-frequency spectrograms or Mel-scaled spectrograms, are a type of spectrogram that uses the Mel scale to warp the frequency axis perceptually. The mel scale better aligns with the human auditory system's frequency perception. To generate a mel spectrogram, the audio signal is divided into frames, and the power spectrum is calculated using techniques like the Fast Fourier Transform (FFT). The resulting power spectrum is then transformed to the mel scale using triangular filter banks. Mel spectrograms are particularly useful in speech analysis tasks, including dysarthria classification. They capture essential spectral information, such as formant frequencies and spectral patterns, while aligning with human perception. Mel spectrograms can be used as image-based features for dysarthria analysis or further processed to extract specific characteristics \cite{ayvaz2022automatic}. Additionally, log Mel spectrograms, obtained by taking the logarithm of the values in the Mel spectrogram, can compress the dynamic range and enhance the visualization and analysis of the mel-frequency content of the audio signal. \cite{joshy2023dysarthria}.

Table 2 summarizes some of the studies in the literature within the collected papers for this review that have utilized image-based features to classify dysarthria based on severity levels.

\subsubsection{Video-based features}
Video-based features in dysarthria analysis involve analyzing the movement of the lips during speech. Lip Movement Analysis aims to extract visual features that provide information about lip shape, lip dynamics, and lip synchronization. By analyzing the visual cues from lip movements, researchers can gain insights into the articulatory aspects of speech production \cite{bandini2018automatic}. However, it is essential to note that classifying the severity of dysarthria solely based on video-based features can be challenging. Dysarthria affects speech mechanisms, including articulation, phonation, resonance, and respiration, which may not be directly observable in video footage. While lip movements can provide valuable information \cite{ackermann1995kinematic}, they may not capture the complete picture of dysarthria symptoms.

It is worth mentioning that research and advancements in computer vision techniques, such as facial landmark tracking and optical flow analysis, continue to improve the estimation of video-based features for dysarthria. These techniques enable more precise extraction and interpretation of lip movements and other facial cues, enhancing the potential for video-based research in dysarthria assessment. 

\subsubsection{Text-related features}
Text-related features  play a crucial role in classifying dysarthria, providing valuable insights into the phonetic characteristics and speech production patterns of individuals with this condition. These features are derived from the analysis of voice signals and focus on the phonetic content of the speech rather than the textual information. Phoneme-level intelligibility serves as a prominent text-related feature, assessing the accuracy of phoneme production by transcribing and comparing the phonetic content of the speech. Additionally, phonetic distance and articulation errors contribute to the classification process, quantifying the dissimilarity between intended and produced phonemes and identifying specific articulation difficulties. By incorporating these text-related features, dysarthria classification models can capture and analyze the phonetic intricacies of speech, enabling improved understanding and identification of different dysarthria subtypes and severity levels. Such as in \cite{xue2023assessing},\cite{xue2023assessing2}, and lexical frequency, phonological neighborhood, word class, and lexical familiarity in \cite{lehner2021impact}.

\subsection{Evaluation}
The evaluation techniques used in machine learning and deep learning models play a crucial role in assessing their performance and effectiveness. These techniques provide insights into the model's predictive capabilities and help practitioners make informed decisions\cite{japkowicz2011evaluating} \cite{powers2020evaluation} \cite{bradley1997use} \cite{hastie2009elements}. These evaluation metrics include \textit{Accuracy} measures the overall correctness of the model's predictions. \textit{Precision and Recall} are commonly used in binary classification tasks, where precision measures the proportion of correctly predicted positive instances and recall measures the proportion of correctly predicted positive instances among all actual positives. The \textit{F1 Score} combines precision and recall into a harmonic mean. \textit{Mean Squared Error (MSE)} is used for regression tasks and measures the average squared difference between predicted and actual values. \textit{Area Under the Curve (AUC)} evaluates binary classifiers by calculating the area under the ROC curve. \textit{Cross-Validation} helps assess model performance across multiple iterations and reduces the risk of overfitting. The \textit{Confusion Matrix} provides a detailed evaluation of the model's performance by showing true positives, true negatives, false positives, and false negatives. These evaluation techniques assist practitioners in assessing model performance, identifying areas for improvement, and comparing different models for a given task.

\section{Discussion}

Let's analyze these categories of features to compare the effectiveness of different features for classifying dysarthria based on severity levels.

Table 1 summarizes studies that have used audio-based features for dysarthria classification based on severity levels. The studies utilize various audio-based features such as fundamental frequency (F0), jitter, shimmer, harmonics-to-noise ratio (HNR), speech rate, pause duration, stress patterns, Mel-Frequency Cepstral Coefficients (MFCCs), Linear Predictive Coding (LPC) coefficients, and formant frequencies (F1, F2, F3, etc.).
Different classification techniques have been used, including Random Forest (RF), Support Vector Machine(SVM), Artificial Neural Network (ANN),  Classification and Regression Tree (CART), Naive Bayes (NB), DNN, CNN, LSTM, Residual Networks (ResNet), multi-layer perception(MLP), Extreme Gradient Boosting (XGBoost), Gaussian Mixture Model(GMM), Probabilistic linear discriminant analysis (PLDA), Hidden Markov Model (HMM), and  k-nearest neighbor(KNN).

It can be seen that most of the researchers' works were evaluated based on standard publicly available datasets such as TORGO \cite{rudzicz2012torgo}, UA Speech \cite{kim2008dysarthric}, Qolt \cite{choi2012dysarthric}, and Numours datasets\cite{menendez1996nemours}, as well as some other locally collected datasets or other less common foreign languages datasets.

Figure 3 shows the distribution of the included papers chosen for this study based on Categorized Features and Dataset Groups. The table  reveals interesting patterns in the distribution of papers across different categorized features and dataset groups in dysarthria classification. Among the featured categories, audio-based techniques demonstrate the highest representation across all dataset groups, including TORGO, UA Speech, Nemours, Qolt, and others. This indicates the prevalent use of audio features in dysarthria classification research across diverse datasets. The prominence of audio-based techniques suggests that researchers prioritize capturing the acoustic characteristics of dysarthric speech for accurate classification.

In contrast, image-based and text-based features have a limited presence, with only a few studies exploring their potential within the UA Speech and other dataset groups. This highlights a potential research gap and suggests the need for further investigation into the utilization of visual information for improved dysarthria classification. Video-based features, on the other hand, are not extensively explored in the selected papers, indicating a less prominent role in dysarthria classification across all dataset groups. We added the mixed category as well, which incorporates multiple feature types, and shows a moderate presence in some dataset groups, underscoring the potential benefits of integrating different modalities to enhance classification performance

The majority of using the audio-based features is due to several advantages. Firstly, they provide a direct measurement of speech by capturing important properties of speech signals, including pitch, intensity, and spectral characteristics. These features are essential in evaluating dysarthria, primarily affecting a person's ability to produce clear and intelligible speech.

Secondly, acoustic features allow for objective and quantitative analysis through signal-processing techniques. This objectivity and quantifiability enhance the consistency and reliability of dysarthria severity assessments. By relying on concrete acoustic properties of speech, these measures provide a more robust evaluation of the condition.

Another benefit is that acoustic analysis is non-invasive and accessible. It can be conducted using standard methods such as recording speech samples with a microphone. This practicality makes acoustic analysis suitable for various settings, including clinics, research studies, and telemedicine. It eliminates the need for invasive procedures or specialized equipment, increasing the ease of implementation.

Furthermore, acoustic features enable longitudinal monitoring of dysarthria progression and treatment outcomes. By analyzing changes in the acoustic properties of speech over time, clinicians and researchers can assess the effectiveness of interventions and track the impact of dysarthria on an individual's communication abilities. This longitudinal perspective provides valuable insights into the management and prognosis of dysarthria \cite{kent2009speech}\cite{lopez2013selection}\cite{luz2017longitudinal}.
\begin{figure}
    \centering
    \includegraphics[width=0.5\textwidth]{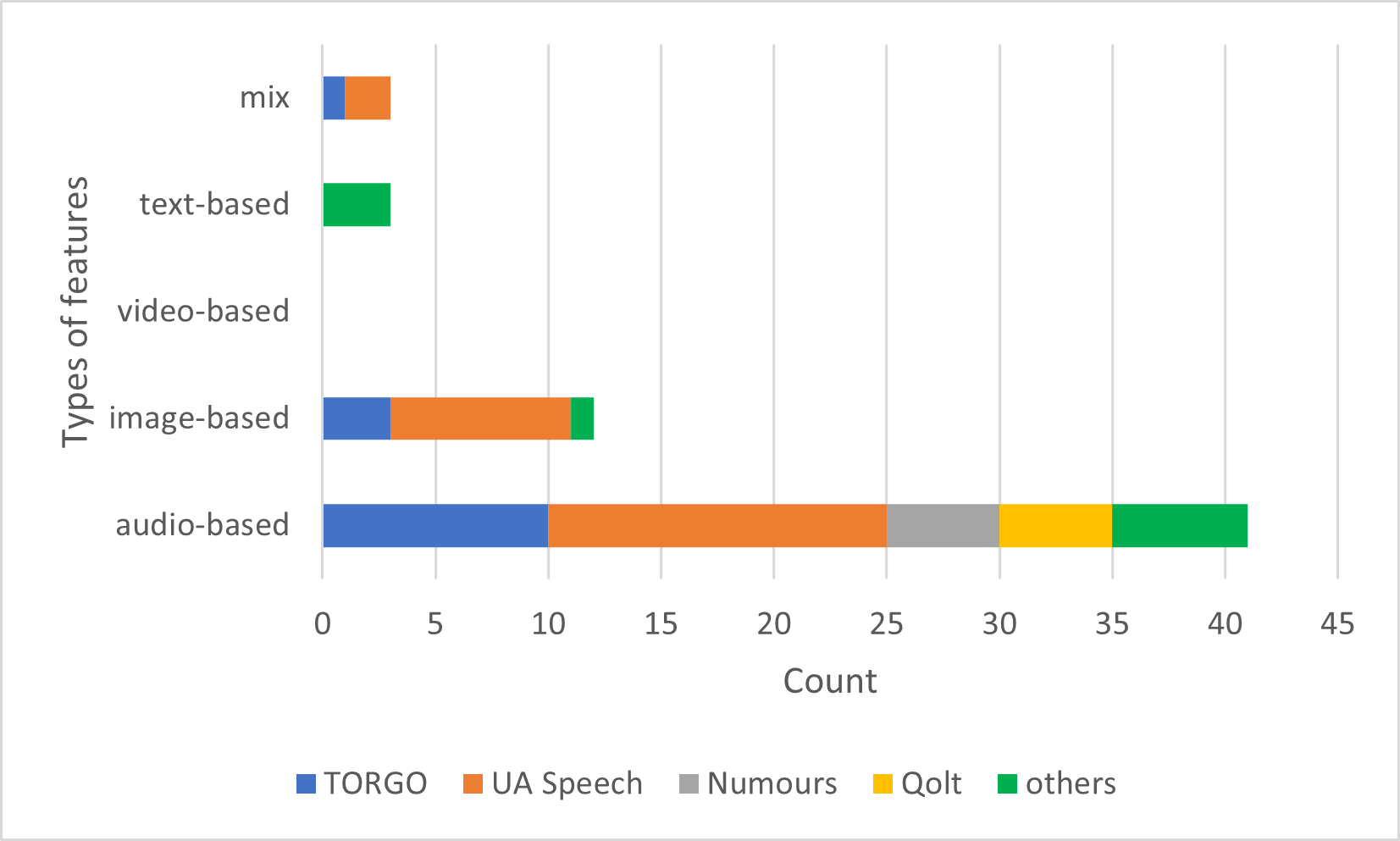}
    \caption{The percentage of each Feature's category}
\end{figure}

While using acoustic features in dysarthria assessment offers several advantages, there are some limitations to consider. Acoustic features partially represent dysarthric speech, as dysarthria encompasses various communication aspects beyond acoustic properties, such as articulation and prosody. Additionally, acoustic analysis lacks the broader context of communication, failing to capture factors like facial expressions or situational cues that influence speech intelligibility.
 
 The generalizability of acoustic features is limited, as they are derived from controlled environments and may not account for real-world variations in acoustic conditions. The acoustic analysis also primarily focuses on objective measures, potentially overlooking subjective experiences and perceptions of individuals with dysarthria. Moreover, distinguishing between dysarthria subtypes solely based on acoustic analysis can be challenging due to shared acoustic characteristics \cite{turner1995influence}\cite{weismer2001acoustic}.
Despite the promising results of the studies in Table 1, it shows some limitations due to the above-mentioned drawbacks of the audio-based features and other reasons related to the type and size of the dataset used for their training and evaluation and the type of the classifiers used.

To overcome these limitations, researchers and clinicians often explore multimodal approaches incorporating other data types, such as images, videos, linguistic features, or perceptual assessments. Integrating multiple modalities can provide a more holistic understanding of dysarthria and improve the accuracy and robustness of the assessment process. As can be seen, some of these attempts are shown in Table 2 for studies that have used image-based features, where most achieved high accuracies around 90\% and above.
Despite the high performance in most studies in Table 2, still have some limitations, including data scarcity, challenges in collecting dysarthric speech data, lower accuracy due to limited datasets, and confinement to isolated utterances.

Some researchers have derived benefits from employing text-based features within the feature types discussed earlier. These features, extracted from the patient's voice signal and converted into text, have demonstrated promising outcomes. However, their usage has significant limitations, rendering exclusive reliance on these types impractical. These limitations are represented by the sound quality, which plays a crucial role in the transcription process. 

Despite sophisticated speech models employed by Speech-to-Text applications, accuracy can be compromised by the type and specifications of the microphones used. Audio input quality might be degraded if users speak too close to or too far from the microphone, which can subsequently impact the transcription's precision. Another fundamental limitation is the presence of background and environmental noise in the audio input. Non-speech sounds can interfere with the audio, leading to less accurate transcriptions. The complexity of transcription can also escalate when multiple speakers speak simultaneously or when there is background speech while the primary user is speaking. Specialized vocabulary presents another challenge. Even though Speech-to-Text models can recognize a wide variety of words, they may stumble upon unique terms or industry-specific jargon not included in the model's vocabulary, resulting in potential transcription errors. Accents and dialects within the same language can pose another difficulty. If a speaker's accent deviates substantially from the norm the model has been trained on; transcription accuracy might decline. Finally, language mismatch can considerably affect transcription accuracy. If the language the user speaks differs from what the Speech-to-Text application expects, the transcription will likely be less accurate. For example, if the system is set to transcribe English, but the user speaks Arabic, the output will likely be flawed \cite{modale2020review}.
\begin{table*}[t!]
\footnotesize
\caption{Summary of the related works used the Audio-based features for classifying dysarthria based on the severity levels}
\label{table_soc}
\centering
\begin{tabular}{|p{2cm}|p{7cm}|p{3cm}|p{3cm}|}
\hline
\textbf{Method} & \textbf{Feature}& \textbf{Classifier}& \textbf{Dataset}\\
\hline 
\citet{chandrashekar2019breathiness}&Breathiness features like HNR&SVM&UA Speech and Kannada Database\\ \hline
\citet{hernandez2020prosody}&prosody, voice quality, MFCC and their combination& RF, SVM, MLP &TORGO, QoLT \\ \hline
\citet{tartarisco2021artificial}&features extracted by VGGish net &ML, TF &Healthy, ataxia subjects  \\ \hline

\citet{joshy2022automated}&Acoustic characteristics, articulatory movements, glottal functioning, and low-dimensional representations&DNN, CNN, gated recurrent units (GRU),and LSTM& UA-Speech and TORGO \\ \hline

\citet{yeo2023automatic}&Hand-crafted acoustic features, as well as raw waveforms &Multi task learning (MLT), SVM, MLP, and XGBoost &QoLT \\ \hline

\citet{karjigi2023speech}&Spectral domain representation features,
Cepstral domain representation features, and Frame-level features &PLDA, and
ANN&UA speech, TORGO \\ \hline

\citet{yeo2021automatic}&MFCCs, Voice Quality Features,	 Prosody Features&SVM,
ANN& QoLT  \\ \hline

\citet{kadi2014automated}&eleven prosodic features selected by LDA&GMM, SVM& Nemours database\\\hline

\citet{al2021classification}&four acoustic features: prosody, spectral, cepstral, and voice quality&SVM, LDA, ANN, CART, NB, RF& Nemours database \\ \hline
   
\citet{bhat2017automatic}&Multi-tapered spectral estimation-based features, Acoustic descriptors for timbre&ANN& UA Speech and TORGO  \\ \hline

 \citet{hernandez2020dysarthria} &Standard Prosodic Features and Rhythm-Based Features&RF, SVM, MLP& QoLT, and TORGO  \\ \hline
 
\citet{kachhi2022teager}&Cepstral Coefficients (TECC, LFCC), and spectral features.&CNN, Light CNN, ResNet& UA Speech \\ \hline
\citet{paja2012automated}& Temporal Features, Spectral Features, Voice Quality Features &Mahalanobis distance&UA Speech  \\ \hline

\citet{yeo2022cross}&Language-independent features from diverse speech dimensions, and language-unique features specific to each language&	XGBoost& TORGO, QoLT, and SSNCE \\ \hline
 
 \citet{kadi2013discriminative}&Prosodic features &	GMM and 
SVM	& Nemours database \\\hline
\citet{vyas2016automatic}&prosodic features: (MFCCs), skewness, and formants.&SVM& UA Speech\\ \hline

\citet{kim2013dysarthric}&10 speech features related to phonetic quality, prosodic quality, and voice quality&SVM	&QoLT\\\hline
\citet{purohit2021weak}&MFCCs&CNN&UA Speech and home service Corpus\\\hline
\citet{kachhi2022significance}&Acoustic/Signal-based Features (SECC, TECC)&CNN,
Light CNN,
ResNet&UA speech  \\ \hline

\citet{mendoza2021acoustic}&10 acoustic features&Discriminant analysis&Dutch speakers and English speakers\\\hline

\citet{gurugubelli2020analytic}&Single Frequency Filter Bank-based Instantaneous Frequency Cepstral Coefficients (SFFB-IFCC)&I-Vector with PLDA Classifier&UA Speech  \\ \hline

\citet{joshy2021automated}&MFCCs and their first two derivatives (a total of 39 features).&SVM, DNN, CNN, LSTM &UA Speech, and TORGO \\ \hline
\citet{narendra2020automatic}&Glottal Parameters and basic acoustic features&SVM&UA Speech\\ \hline
\citet{gillespie2017cross}&Spectral, prosodic, Teager Energy Operator (TEO), and glottal waveform features& SVM &UA Speech \\ \hline
\citet{kadi2016fully}&traditional acoustic features like MFCC and auditory-based cues. &GMM, SVM, and hybrid GMM/SVM & the Nemours database  and  Torgo \\ \hline
 
 \citet{dahmani2014relevance}&Rhythm Metrics	Various rhythm metrics, such as SRVC (speaking rate of VC segments)&SVM& Nemours corpus\\\hline
 \citet{joshy2023statistical}&prosody, articulation, phonation, and glottal &SVM, NB, kNN, and RF&UA Speech and TORGO  \\\hline
 
 \citet{javanmardi2023wav2vec}&wav2vec features &SVM&UA Speech\\\hline
\citet{demino2011assessing}&Frame-level features with global statistics &Forward selection method (FSM)and boosting algorithm&Dataset of 39 dysarthria patients\\\hline
\end{tabular}
\end{table*}

Some approaches may combine audio with image or video features to capture a more comprehensive representation of dysarthric speech, such as in \cite{tong2020automatic} where audio-visual joint features are used as input to the CNN and evaluated on UA Speech dataset with an accuracy up to 99.5\%. Despite the high performance of their work, their techniques face a few limitations related to manual data pre-processing and high computational power requirements. In \cite{kachhi2022continuous}, the authors applied several types of features rather than mixing them as one input to the classifier to check which type is more effective in classifying dysarthria based on severity levels, where they used Generalized Morse Wavelet (GMW)-based scalogram features (for low-frequency areas) and Mel spectrogram-based features with CNN. They evaluated their work on the UA Speech dataset and achieved the highest accuracy using Scalgram features at 95.17\%. Another work where the same has been done in \cite{kumar2022towards} where several experiments on MFCC, the audio combination of features including( ZCR (Zero Crossing Rate), Spectral centroid, spectral roll-off) and Mel spectrogram using KNN, and SVM classifiers and evaluated their work on Torgo with an accuracy of 95\% using SVM for mel spectrogram.

It's important to note that the choice of features depends on the specific goals and requirements of the dysarthria severity assessment task. Different features may carry different levels of information and may be more or less suitable for various applications. Researchers and practitioners often select and combine features based on their relevance and effectiveness in capturing the characteristics of dysarthric speech. 

\subsection{Discussion of Research Question 1}

Several types of features were used across the studies:
The most common type of feature is the Acoustic feature. These include MFCCs (Mel Frequency Cepstral Coefficients), voice quality, and spectral and prosodic features (such as pitch and rhythm). 
In a general analysis, most studies used a combination of these features.
Some studies focused on features Domain-specific features of speech disorders, such as breathiness features\cite{chandrashekar2019breathiness} and parameters related to glottal function\cite{joshy2022automated}. Hand-crafted and raw waveform features were used by one study that combined these two approaches\cite {yeo2023automatic}.
As shown in Figure 4, the mean accuracy obtained by the most utilized classification techniques representing the highest accuracy for each group of the datasets based on audio features reached up to $90s$. 
In specific analysis, Complex, multidimensional feature sets (a combination of features) tend to perform better and are associated with high accuracy rates. For example and referring to individual performance values extracted from the original papers, Ref.\cite{joshy2022automated} using MFCCs, constant-Q cepstral coefficients, prosody, articulation, phonation, and glottal functioning achieved 93.97\% accuracy. Another example is the study that achieved 99.9\% with UA Speech dataset \cite{karjigi2023speech} used a mix of spectral, cepstral, and frame-level features. While combining multiple features and techniques can result in high accuracies, it can also introduce limitations such as increased computational time \cite{joshy2022automated} and difficulties in interpreting feature importance \cite{yeo2021automatic}. 

Prosodic features, including articulation, often demonstrate high performance in dysarthria classification. These features capture essential aspects of speech production, such as rhythm, intonation, and speech clarity, which are crucial for assessing dysarthria severity. In \citet{al2021classification} reported that prosodic features are the best for classifying the mild cases (which are better classified compared to severe and moderate levels due to having more common features among speakers), and the moderate cases over the spectral features, which achieved remarkable results for classifying severe cases. Cepstral features are less effective for classifying severity levels. This is also approved by \cite{kadi2014automated} and \citet{joshy2023statistical}. 

 Some other techniques explored individual features for the classification purposes of ineligibility levels or dysarthria severity levels and discovered some significant features which can achieve a high accuracy rate, such as in \cite{chandrashekar2019breathiness}, which relied on several types of breathiness features represented by Jitter, Shimmer, Harmonic-to-Noise Ratio, Harmonic Energy, Harmonic Energy of Residue, Harmonic-to-Signal Ratio, and Glottal-to-Noise Excitation Ratio and reached 96\%. In their work, they discovered that Harmonic-related features could distinguish intelligibility levels and achieve the highest accuracy compared to other features but struggled with mild dysarthria. Also, in \cite{purohit2021weak}, where the authors rely only on the MFCC features set and achieved up to 99\% accuracy rate after experiencing several trails with different utterances lengths, 50, 200, and 300 and found that increasing the length of utterances will lead to better performance. A similar attempt relying on MFCC is in \cite{lee2019assessment}. 

Concerning image-based features, the Mel spectrograms and their log or derivatives are the primary types of features and are primarily evaluated on TORGO and UA Speech datasets, as shown in Figure 5. This is a benefit of using only one type to save the preparation and extraction of too many features, as the network is responsible for achieving this task.

The mean accuracies for classifying dysarthria based on severity levels varied across different datasets. Among the selected datasets, the TORGO dataset exhibited moderate mean accuracies ranging from 73.99\% (KNN) to 92.47\% (DL), with a notable performance from CNN (88.15\%). The UA Speech dataset showed higher mean accuracies, with MLP achieving 98.17\% accuracy, followed closely by LSTM (96.94\%) and CNN (96.23\%). The Qolt dataset had lower mean accuracies, with SVM and RF achieving approximately 70\% accuracy. Notably, the Numours dataset demonstrated high mean accuracies of 94.45\% (SVM) and 95.8\% (RF). The Others dataset had an outstanding performance, with SVM and CNN achieving mean accuracies of 96\% and 91.8\%, respectively.

These mean accuracies provide insights into the effectiveness of classifiers for different datasets in the context of severity-based dysarthria classification. It is important to note that the performance of classifiers can vary depending on the dataset characteristics, such as the number of samples, data quality, and diversity of dysarthria types.

\begin{table*}[t!]
\caption{Summary of the related works used Image-based features for classifying dysarthria based on the severity levels}

\centering
\begin{tabular}{|p{2cm}|p{5cm}|p{4cm}|p{3cm}|}
\hline
\textbf{Method} & \textbf{Feature}& \textbf{Classifier}& \textbf{Dataset}\\
\hline 

\citet{joshy2023dysarthria}&Mel spectrograms&Multi-head attention mechanism (MHA), Multi-task learning, and ResNet&UA Speech \\\hline

\citet{joshy2023dysarthria2}&Mel spectrograms&Squeeze-and-excitation (SE) networks&UA Speech\\\hline

\citet{chandrashekar2020investigation}&perceptually enhanced Fourier transform spectrograms and constant-Q transform (CQT) spectrograms& CNN& UA speech
and TORGO  \\\hline

\citet{suhas2020speech}& log Mel spectrograms& CNN& 60 patients from each amyotrophic lateral sclerosis (ALS), Parkinson's disease (PD), and healthy controls\\\hline
\citet{fernandez2020attention}&Log-mel spectrograms&	LSTM networks with Attention mechanism	&	UA Speech  \\ \hline
\citet{montalbo4442941dysarnet}& 2D image spectrograms&DySARNet, a densely squeezed-and-excited attention-gated residual neural network&TORGO and UA-Speech\\ \hline
\citet{gupta2021residual}&Spectrograms of short speech segments &ResNet (CNNs, GMM, and Light CNNs for comparison)&UA Speech \\ \hline
\citet{chandrashekar2019spectro}&Spectrogram and its derivatives&CNN&UA Speech  and TORGO \\\hline
\end{tabular}
\end{table*}

\subsection{Discussion of Research Question 2}

Regarding the classification techniques, advanced algorithms such as Support Vector Machines (SVM), Discriminant analysis, and deep learning models (DNN, CNN, LSTM) were frequently seen and generally delivered a strong performance. For example, SVM was widely utilized across various studies, including \cite{hernandez2020prosody}, \cite{al2021classification}, \cite{hernandez2020dysarthria}, \cite{yeo2021automatic}, \cite{yeo2022cross}, \cite{yeo2023automatic}, \cite{kim2013dysarthric}, \cite{kadi2014automated}, \cite{narendra2020automatic}, \cite{kadi2016fully}, \cite{joshy2023statistical}, and \cite{tartarisco2021artificial}. 

ANN-based models, seen in studies such as \cite{hernandez2020prosody}, \cite{al2021classification}, \cite{bhat2017automatic}, \cite{hernandez2020dysarthria}, \cite{yeo2021automatic}, and \cite{karjigi2023speech}. RF showed in \cite{hernandez2020prosody}, \cite{al2021classification},  \cite{hernandez2020dysarthria}, and \cite{joshy2023statistical}, and Deep learning techniques (CNN, DNN, LSTM, etc.), applied in studies \cite{joshy2022automated}, \cite{kachhi2022teager}, \cite{joshy2021automated}, \cite{javanmardi2023wav2vec}, yielded high accuracies, typically above 90\%.

Newer techniques like self-supervised learning with Wav2vec 2.0 XLS-R, as seen in \cite{yeo2023automatic}, didn't perform as well, achieving only a 65.52\% accuracy rate. This highlights the potential challenges in adapting these models for dysarthria severity classification.

From Figure 4, audio-based classification, SVM and RF demonstrated relatively high mean accuracies across most datasets. In the TORGO dataset, SVM achieved a mean accuracy of 77.28\%, while RF achieved 79.17\% of mean accuracy. In the UA Speech dataset, both SVM and RF achieved accuracies above 80\%, with SVM reaching 80.25\% and RF reaching 82.69\%. However, in the Qolt dataset, the mean accuracies for SVM and RF dropped to 70\% and 70\%, respectively. It's important to note that RF achieved a high accuracy of 95.8\% in the Numours dataset, while SVM achieved a remarkable accuracy of 96\% in the Others dataset.

For image-based classification in Figure 5, CNN and LSTM were the prominent techniques, as deep learning techniques is a direct way to extract features and classify them into several severity levels using their different layers functions. In the TORGO dataset, CNN achieved an accuracy of 88.15\%, while LSTM achieved 99.2\% accuracy. In the UA Speech dataset, CNN demonstrated high accuracy of 96.23\%, and LSTM achieved 96.94\% accuracy. Notably, ResNet only yields  accuracy for the UA Speech  datasets around 97\%.

These results suggest that for audio-based classification, SVM and RF show consistent performance across various datasets. Despite the effectiveness of using SVM, it is reported in \citet{kadi2016fully}that it is not adequate to process utterances with diverse time lengths and needs to unify the time lengths for all speech utterances. For image-based classification, CNN and LSTM exhibit high accuracies. However, further investigation and evaluation are needed to determine these techniques' generalizability and performance on more extensive and diverse dysarthria datasets.
 
The mean accuracies presented in the tables were obtained by calculating the average of the highest accuracies reported in the original papers for each dataset group. To ensure accuracy and reliability, only the highest accuracy value from each paper was considered. This approach allows for a comprehensive analysis of the performance of the classification techniques across different datasets, taking into account the number of papers that reported the respective accuracies. By aggregating the highest accuracies, we provide a representative measure of the overall performance of the techniques for each dataset group, considering the number of papers that contributed to calculating the mean accuracies.

\begin{figure}
    \centering
    \includegraphics[width=0.5\textwidth]{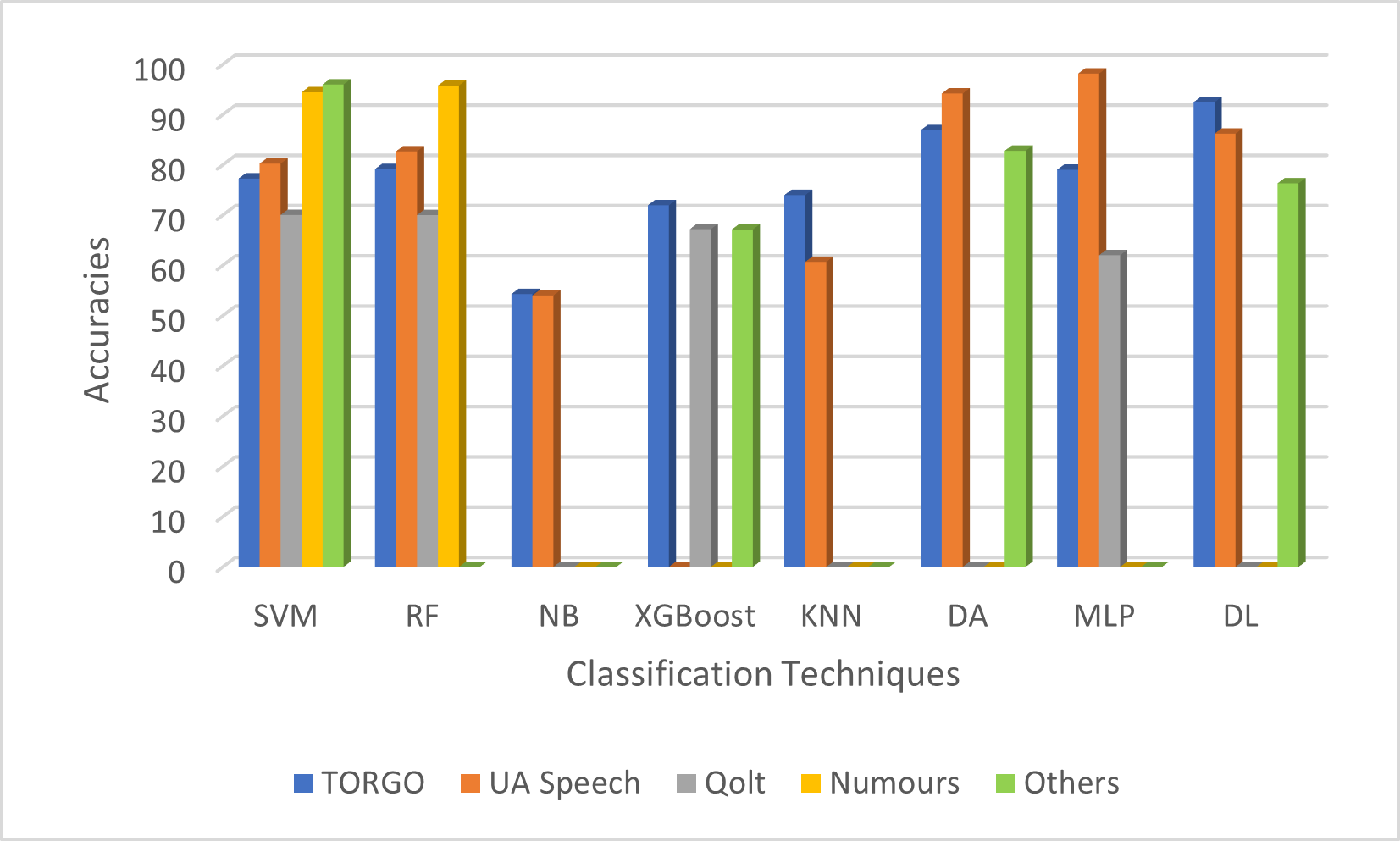}
    \caption{Mean Accuracy for Audio-based Features for common datasets}
\end{figure}
\begin{figure}
    \centering
    \includegraphics[width=0.5\textwidth]{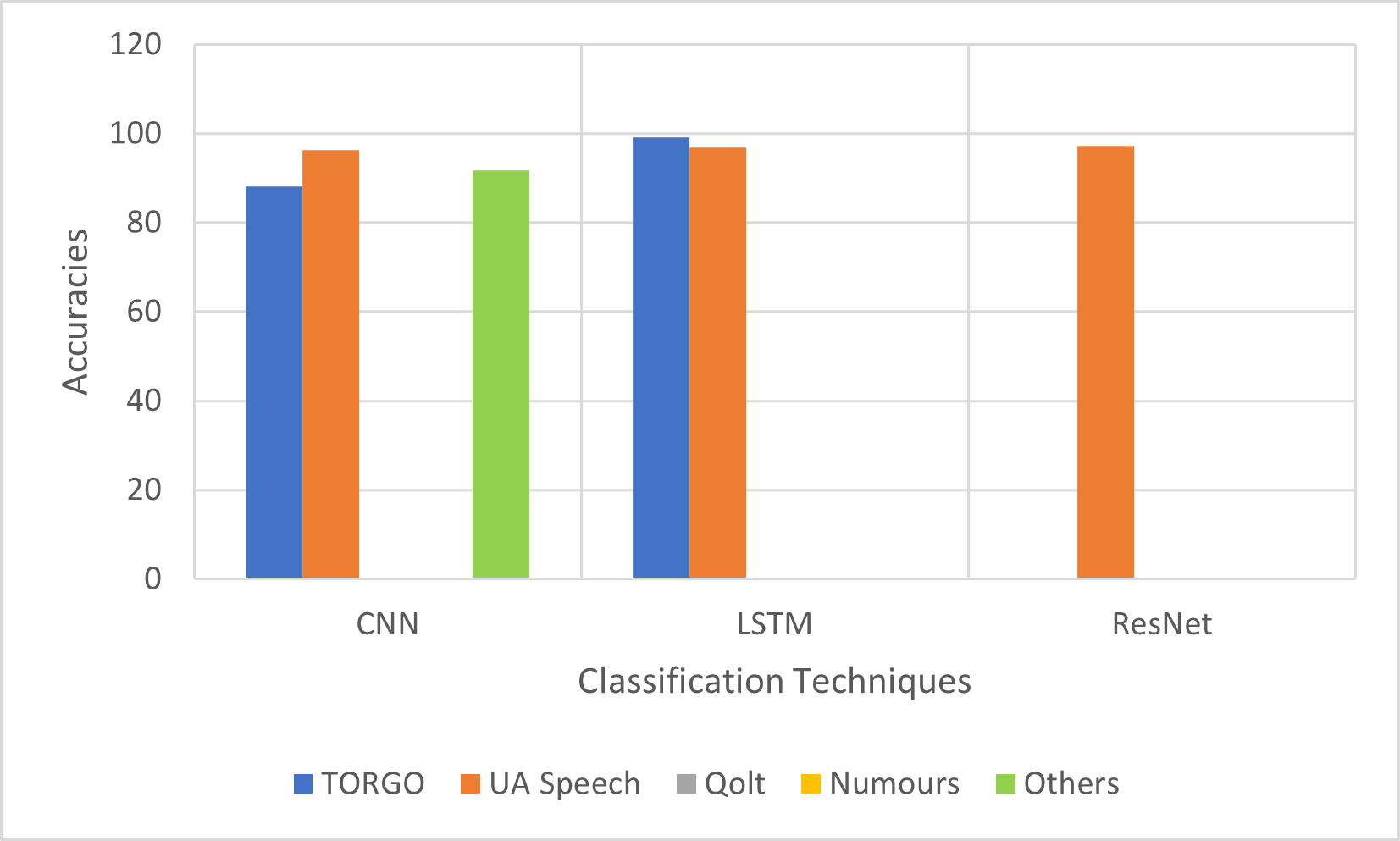}
    \caption{Mean Accuracy for Image-based Features for common datasets}
    \label{fig:enter-label}
\end{figure}

\section{Limitations}

The audio-based features used in dysarthria classification have certain limitations, which can be compared to image-based features. By analyzing these limitations, we can gain insights into the techniques and accuracies associated with each feature type.

Regarding audio-based features, one limitation mentioned in \cite{chandrashekar2019breathiness} is the struggle to accurately classify cases with mild dysarthria. This implies that audio-based features might be more effective in capturing severe dysarthria characteristics, while their discriminatory power decreases with milder cases. Additionally, there are difficulties in interpreting the features and handling data diversity \cite{hernandez2020prosody}. This suggests that some audio-based features may lack clear interpretability and may not be robust enough to handle the diverse dysarthria characteristics. Furthermore, there are limitations related to small datasets and the focus on a single type of dysarthria \cite{al2021classification}, which can hinder the generalizability of the findings.

On the other hand, image-based features also have their limitations. For instance, inter-subject variability and gender bias can impact the performance of image-based models \cite{joshy2023dysarthria2}. These limitations suggest that certain image-based features might be more influenced by individual differences, making them less reliable for general dysarthria classification. Another limitation is the computational time required for specific image-based techniques \cite{chandrashekar2020investigation}, which can hinder their real-time application.

When comparing the techniques and accuracies, it is essential to note that the performance varies across different studies and datasets. Audio-based features have reported accuracies ranging from 40.41\% to 95.80\% \cite{al2021classification}, whereas image-based features have achieved accuracies ranging from 72\% to 99.20\% \cite{chandrashekar2020investigation, montalbo4442941dysarnet}. These variations highlight the influence of the chosen techniques and datasets on the achieved accuracies.

In summary, both audio-based and image-based features are limited in dysarthria classification. Audio-based features struggle with mild dysarthria cases, interpretation difficulties, and limited datasets, while image-based features face challenges related to inter-subject variability, gender bias, and computational time. These limitations influence the techniques used with these features and the resulting accuracies, emphasizing the need to carefully consider the feature type, technique selection, and dataset characteristics in dysarthria classification research.

\section{Suggested solutions}
To overcome the limitation cited above, some analyses are performed using computer vision features that can improve the performance of each method.

Solving the issue of larger data size: Instead of utilizing only the standard machine learning techniques or the deep learning techniques which require large amounts of the dataset, we suggest using rule-based models such as Adaptive Neuro-Fuzzy Inference System (ANFIS), which outperform the other methods in terms of the necessity to small data size for training either in terms of features or samples, the more interpretable model explicitly compared to deep learning, adaptable and generalized model due to the generated rules which can be adjusted based on the specific application and the used data nature and the non-linearity and noise handling \cite{akkocc2012empirical}\cite{subasi2007application}.

Some new features are suggested to be used by researchers in this field inclduing Nonlinear Dynamical Features, Vocal Tract Resonance Estimation, Source-Filter Separation Features, Microprosody Analysis, and Articulatory Kinematics.

 \textbf{Nonlinear Dynamical Features}: Dysarthria affects the coordination and control of speech production, resulting in altered dynamics. Nonlinear dynamical features, such as recurrence plots(can help visualize temporal patterns in speech features that could indicate varying levels of severity \cite{geman2011data}), Variations in the fractal dimension of speech signals could distinguish between different severity levels of dysarthria \cite{accardo1998algorithm}, or entropy measures (e.g., permutation entropy, sample entropy) (lower entropy indicates a less complex, more predictable one, and dysarthria's speech signals may become less complicated due to muscle weakness and coordination difficulties, can capture the underlying complexity and temporal organization of dysarthric speech signals \cite{rudzicz2010towards}). We know these features have not been utilized to classify dysarthria. Still, it is used with speech analysis, such as in \cite{jackson2016recurrence}, and it is helpful to be calculated and combined with video-based features.

         \textbf{Vocal Tract Resonance Estimation}: Investigate techniques for estimating the vocal tract resonances directly from the speech signal. This could involve formant tracking, subspace-based analysis, or model-based estimation. This feature type was utilized for classifying dysarthria subtypes in \cite{leveque2022acoustic}. These features alone may not be sufficient for classifying dysarthria based on severity levels. They primarily capture information about the resonant characteristics of the vocal tract, which can be influenced by various factors such as vocal fold movement, articulatory precision, and vocal tract shape. So, combining them with other acoustic features may help build a successful classification model.

        \textbf{Source-Filter Separation Features}: Dysarthria can affect the speaker's source (glottal excitation) and filter (vocal tract) components. Extracting features that specifically focus on the characteristics of the glottal source, such as measures related to the glottal flow derivative or instantaneous frequency, in combination with standard spectral features, can provide a comprehensive representation for dysarthria classification. These features have been utilized in \cite{yue2022raw} for automatic dysarthric speech recognition and evaluated on the TORGO dataset, which contains diverse severity samples. Still, there is no specific classification model based on these features.

         \textbf{Microprosody Analysis}: Investigate features that capture fine-grained temporal variations within short speech segments, called microporosity. This could involve analyzing pitch fluctuations, intensity modulations, or spectral changes at very short time scales (e.g., using wavelet transforms or time-frequency representations). These features can potentially capture subtle dysarthric speech dynamics. These types of features used mainly with speech synthesis and its variations were also explored in many studies such as \cite{vainio1998modeling}.

        \textbf{Articulatory Kinematics}: Consider exploring articulatory and kinematic features, such as lip or jaw motion tracking. Using techniques like optical motion capture or electromagnetic articulography,  features can be extracted that describe articulatory gestures' spatial and temporal characteristics, which may be informative for dysarthria classification. These features have a notable correlation with the severity levels of dysarthria as discussed in \cite{lee2018articulatory}, and they have been used to classify vocal segments into control, PD, and ALS, but still no specific work on classifying dysarthria based on severity utilizing this type of features.

\section{Conclusion}
Accurately classifying dysarthria based on severity levels is crucial for clinicians and researchers, as it provides valuable insights into the disorder's impact on communication abilities. This classification enables tailored treatment strategies, objective assessment measures, and progress tracking, ultimately improving the quality of life for individuals with dysarthria. This review aimed to analyze the classification of dysarthria based on severity levels, focusing on the effectiveness of different features and artificial intelligence techniques. The use of audio-based features, mainly acoustic analysis, has been the predominant approach in dysarthria classification. However, limitations such as data scarcity and challenges in real-time implementation have been observed. To overcome these limitations, multimodal systems integrating other data types have been explored. Complex, multidimensional feature sets tend to perform better but can introduce challenges such as increased computational time. The choice of AI techniques does not directly correlate with higher accuracy, as both traditional and advanced techniques have shown varying performance. In conclusion, integrating multiple modalities, exploring new feature types, and carefully selecting AI techniques can improve the accuracy and efficiency of dysarthria classification systems. Future research should address the identified limitations and refine the classification methods for enhanced clinical utility.

\section* {Declaration of competing interest}
{The authors declare no conflicts of interest.}

\section*{Acknowledgements}

This publication was made by International Research Collaboration Co-Fund (IRCC) Cycle 06
(2023-2024) No. IRCC-2023-223 from the Qatar National Research Fund (a member of the Qatar Foundation). The statements made herein are solely the responsibility of the authors.

\bibliographystyle{elsarticle-num-names}

\begin{thebibliography}{108}
\expandafter\ifx\csname natexlab\endcsname\relax\def\natexlab#1{#1}\fi
\providecommand{\url}[1]{\texttt{#1}}
\providecommand{\href}[2]{#2}
\providecommand{\path}[1]{#1}
\providecommand{\DOIprefix}{doi:}
\providecommand{\ArXivprefix}{arXiv:}
\providecommand{\URLprefix}{URL: }
\providecommand{\Pubmedprefix}{pmid:}
\providecommand{\doi}[1]{\href{http://dx.doi.org/#1}{\path{#1}}}
\providecommand{\Pubmed}[1]{\href{pmid:#1}{\path{#1}}}
\providecommand{\bibinfo}[2]{#2}
\ifx\xfnm\relax \def\xfnm[#1]{\unskip,\space#1}\fi
\bibitem[{Shriberg et~al.(2019)Shriberg, Strand, Jakielski, and Mabie}]{shriberg2019estimates}
\bibinfo{author}{L.~D. Shriberg}, \bibinfo{author}{E.~A. Strand}, \bibinfo{author}{K.~J. Jakielski}, \bibinfo{author}{H.~L. Mabie},
\newblock \bibinfo{title}{Estimates of the prevalence of speech and motor speech disorders in persons with complex neurodevelopmental disorders},
\newblock \bibinfo{journal}{Clinical Linguistics \& Phonetics} \bibinfo{volume}{33} (\bibinfo{year}{2019}) \bibinfo{pages}{707--736}.
\bibitem[{Duffy(2019)}]{duffy2019motor}
\bibinfo{author}{J.~R. Duffy}, \bibinfo{title}{Motor speech disorders e-book: Substrates, differential diagnosis, and management}, \bibinfo{publisher}{Elsevier Health Sciences}, \bibinfo{year}{2019}.
\bibitem[{Kent and Rosenbek(1983)}]{kent1983acoustic}
\bibinfo{author}{R.~D. Kent}, \bibinfo{author}{J.~C. Rosenbek},
\newblock \bibinfo{title}{Acoustic patterns of apraxia of speech},
\newblock \bibinfo{journal}{Journal of Speech, Language, and Hearing Research} \bibinfo{volume}{26} (\bibinfo{year}{1983}) \bibinfo{pages}{231--249}.
\bibitem[{Laganaro et~al.(2021)Laganaro, Fougeron, Pernon, Lev{\^e}que, Borel, Fournet, Catalano~Chiuv{\'e}, Lopez, Trouville, M{\'e}nard et~al.}]{laganaro2021sensitivity}
\bibinfo{author}{M.~Laganaro}, \bibinfo{author}{C.~Fougeron}, \bibinfo{author}{M.~Pernon}, \bibinfo{author}{N.~Lev{\^e}que}, \bibinfo{author}{S.~Borel}, \bibinfo{author}{M.~Fournet}, \bibinfo{author}{S.~Catalano~Chiuv{\'e}}, \bibinfo{author}{U.~Lopez}, \bibinfo{author}{R.~Trouville}, \bibinfo{author}{L.~M{\'e}nard}, et~al.,
\newblock \bibinfo{title}{Sensitivity and specificity of an acoustic-and perceptual-based tool for assessing motor speech disorders in french: The monpage-screening protocol},
\newblock \bibinfo{journal}{Clinical Linguistics \& Phonetics} \bibinfo{volume}{35} (\bibinfo{year}{2021}) \bibinfo{pages}{1060--1075}.
\bibitem[{Chandrashekar et~al.(2019)Chandrashekar, Karjigi, and Sreedevi}]{chandrashekar2019breathiness}
\bibinfo{author}{H.~Chandrashekar}, \bibinfo{author}{V.~Karjigi}, \bibinfo{author}{N.~Sreedevi},
\newblock \bibinfo{title}{Breathiness indices for classification of dysarthria based on type and speech intelligibility},
\newblock in: \bibinfo{booktitle}{2019 International Conference on Wireless Communications Signal Processing and Networking (WiSPNET)}, \bibinfo{organization}{IEEE}, \bibinfo{year}{2019}, pp. \bibinfo{pages}{266--270}.
\bibitem[{Tong(2020)}]{tong2020automatic}
\bibinfo{author}{H.~Tong}, \bibinfo{title}{Automatic assessment of dysarthric severity level using audio-video cross-modal approach in deep learning}, Master's thesis, \bibinfo{year}{2020}.
\bibitem[{Noffs et~al.(2018)Noffs, Perera, Kolbe, Shanahan, Boonstra, Evans, Butzkueven, van~der Walt, and Vogel}]{noffs2018speech}
\bibinfo{author}{G.~Noffs}, \bibinfo{author}{T.~Perera}, \bibinfo{author}{S.~C. Kolbe}, \bibinfo{author}{C.~J. Shanahan}, \bibinfo{author}{F.~M. Boonstra}, \bibinfo{author}{A.~Evans}, \bibinfo{author}{H.~Butzkueven}, \bibinfo{author}{A.~van~der Walt}, \bibinfo{author}{A.~P. Vogel},
\newblock \bibinfo{title}{What speech can tell us: A systematic review of dysarthria characteristics in multiple sclerosis},
\newblock \bibinfo{journal}{Autoimmunity reviews} \bibinfo{volume}{17} (\bibinfo{year}{2018}) \bibinfo{pages}{1202--1209}.
\bibitem[{Gandhi et~al.(2020)Gandhi, Tobin, Vongphakdi, Copley, and Watter}]{gandhi2020scoping}
\bibinfo{author}{P.~Gandhi}, \bibinfo{author}{S.~Tobin}, \bibinfo{author}{M.~Vongphakdi}, \bibinfo{author}{A.~Copley}, \bibinfo{author}{K.~Watter},
\newblock \bibinfo{title}{A scoping review of interventions for adults with dysarthria following traumatic brain injury},
\newblock \bibinfo{journal}{Brain injury} \bibinfo{volume}{34} (\bibinfo{year}{2020}) \bibinfo{pages}{466--479}.
\bibitem[{Mackenzie(2011)}]{mackenzie2011dysarthria}
\bibinfo{author}{C.~Mackenzie},
\newblock \bibinfo{title}{Dysarthria in stroke: a narrative review of its description and the outcome of intervention},
\newblock \bibinfo{journal}{International journal of speech-language pathology} \bibinfo{volume}{13} (\bibinfo{year}{2011}) \bibinfo{pages}{125--136}.
\bibitem[{Lee et~al.(2021)Lee, Madhavan, Krajewski, and Lingenfelter}]{lee2021assessment}
\bibinfo{author}{J.~Lee}, \bibinfo{author}{A.~Madhavan}, \bibinfo{author}{E.~Krajewski}, \bibinfo{author}{S.~Lingenfelter},
\newblock \bibinfo{title}{Assessment of dysarthria and dysphagia in patients with amyotrophic lateral sclerosis: Review of the current evidence},
\newblock \bibinfo{journal}{Muscle \& Nerve} \bibinfo{volume}{64} (\bibinfo{year}{2021}) \bibinfo{pages}{520--531}.
\bibitem[{Finch et~al.(2020)Finch, Rumbach, and Park}]{finch2020speech}
\bibinfo{author}{E.~Finch}, \bibinfo{author}{A.~F. Rumbach}, \bibinfo{author}{S.~Park},
\newblock \bibinfo{title}{Speech pathology management of non-progressive dysarthria: a systematic review of the literature},
\newblock \bibinfo{journal}{Disability and Rehabilitation} \bibinfo{volume}{42} (\bibinfo{year}{2020}) \bibinfo{pages}{296--306}.
\bibitem[{Whillans et~al.(2022)Whillans, Lawrie, Cardell, Kelly, and Wenke}]{whillans2022systematic}
\bibinfo{author}{C.~Whillans}, \bibinfo{author}{M.~Lawrie}, \bibinfo{author}{E.~A. Cardell}, \bibinfo{author}{C.~Kelly}, \bibinfo{author}{R.~Wenke},
\newblock \bibinfo{title}{A systematic review of group intervention for acquired dysarthria in adults},
\newblock \bibinfo{journal}{Disability and Rehabilitation} \bibinfo{volume}{44} (\bibinfo{year}{2022}) \bibinfo{pages}{3002--3018}.
\bibitem[{Park and Lee(2020)}]{park2020effect}
\bibinfo{author}{Y.-J. Park}, \bibinfo{author}{J.-M. Lee},
\newblock \bibinfo{title}{Effect of acupuncture intervention and manipulation types on poststroke dysarthria: a systematic review and meta-analysis},
\newblock \bibinfo{journal}{Evidence-Based Complementary and Alternative Medicine} \bibinfo{volume}{2020} (\bibinfo{year}{2020}).
\bibitem[{Shanmugam and Marimuthu(2021)}]{shanmugam2021critical}
\bibinfo{author}{A.~K. Shanmugam}, \bibinfo{author}{R.~Marimuthu},
\newblock \bibinfo{title}{A critical analysis and review of assistive technology: advancements, laws, and impact on improving the rehabilitation of dysarthric patients},
\newblock \bibinfo{journal}{Handbook of Decision Support Systems for Neurological Disorders}  (\bibinfo{year}{2021}) \bibinfo{pages}{263--281}.
\bibitem[{Rowe et~al.(2022)Rowe, Gutz, Maffei, Tomanek, and Green}]{rowe2022characterizing}
\bibinfo{author}{H.~P. Rowe}, \bibinfo{author}{S.~E. Gutz}, \bibinfo{author}{M.~F. Maffei}, \bibinfo{author}{K.~Tomanek}, \bibinfo{author}{J.~R. Green},
\newblock \bibinfo{title}{Characterizing dysarthria diversity for automatic speech recognition: A tutorial from the clinical perspective},
\newblock \bibinfo{journal}{Frontiers in Computer Science}  (\bibinfo{year}{2022}) \bibinfo{pages}{43}.
\bibitem[{Chiaramonte and Vecchio(2021)}]{chiaramonte2021systematic}
\bibinfo{author}{R.~Chiaramonte}, \bibinfo{author}{M.~Vecchio},
\newblock \bibinfo{title}{A systematic review of measures of dysarthria severity in stroke patients},
\newblock \bibinfo{journal}{Pm\&r} \bibinfo{volume}{13} (\bibinfo{year}{2021}) \bibinfo{pages}{314--324}.
\bibitem[{Palmer and Enderby(2007)}]{palmer2007methods}
\bibinfo{author}{R.~Palmer}, \bibinfo{author}{P.~Enderby},
\newblock \bibinfo{title}{Methods of speech therapy treatment for stable dysarthria: A review},
\newblock \bibinfo{journal}{Advances in Speech Language Pathology} \bibinfo{volume}{9} (\bibinfo{year}{2007}) \bibinfo{pages}{140--153}.
\bibitem[{Huang et~al.(2021)Huang, Hall, Watson, and Shahamiri}]{huang2021review}
\bibinfo{author}{A.~Huang}, \bibinfo{author}{K.~Hall}, \bibinfo{author}{C.~Watson}, \bibinfo{author}{S.~R. Shahamiri},
\newblock \bibinfo{title}{A review of automated intelligibility assessment for dysarthric speakers},
\newblock in: \bibinfo{booktitle}{2021 International Conference on Speech Technology and Human-Computer Dialogue (SpeD)}, \bibinfo{organization}{IEEE}, \bibinfo{year}{2021}, pp. \bibinfo{pages}{19--24}.
\bibitem[{Stipancic et~al.(2021)Stipancic, Palmer, Rowe, Yunusova, Berry, and Green}]{stipancic2021you}
\bibinfo{author}{K.~L. Stipancic}, \bibinfo{author}{K.~M. Palmer}, \bibinfo{author}{H.~P. Rowe}, \bibinfo{author}{Y.~Yunusova}, \bibinfo{author}{J.~D. Berry}, \bibinfo{author}{J.~R. Green},
\newblock \bibinfo{title}{“you say severe, i say mild”: Toward an empirical classification of dysarthria severity},
\newblock \bibinfo{journal}{Journal of Speech, Language, and Hearing Research} \bibinfo{volume}{64} (\bibinfo{year}{2021}) \bibinfo{pages}{4718--4735}.
\bibitem[{McNeil et~al.(2017)McNeil, Ballard, Duffy, Wambaugh, van Lieshout, Maassen, and Terband}]{mcneil2017apraxia}
\bibinfo{author}{M.~R. McNeil}, \bibinfo{author}{K.~J. Ballard}, \bibinfo{author}{J.~R. Duffy}, \bibinfo{author}{J.~Wambaugh}, \bibinfo{author}{P.~van Lieshout}, \bibinfo{author}{B.~Maassen}, \bibinfo{author}{H.~Terband},
\newblock \bibinfo{title}{Apraxia of speech theory, assessment, differential diagnosis, and treatment: Past, present, and future},
\newblock \bibinfo{journal}{Speech motor control in normal and disordered speech: Future developments in theory and methodology}  (\bibinfo{year}{2017}) \bibinfo{pages}{195--221}.
\bibitem[{Barkmeier-Kraemer and Clark(2017)}]{barkmeier2017speech}
\bibinfo{author}{J.~M. Barkmeier-Kraemer}, \bibinfo{author}{H.~M. Clark},
\newblock \bibinfo{title}{Speech--language pathology evaluation and management of hyperkinetic disorders affecting speech and swallowing function},
\newblock \bibinfo{journal}{Tremor and Other Hyperkinetic Movements} \bibinfo{volume}{7} (\bibinfo{year}{2017}).
\bibitem[{Enderby(1980)}]{enderby1980frenchay}
\bibinfo{author}{P.~Enderby},
\newblock \bibinfo{title}{Frenchay dysarthria assessment},
\newblock \bibinfo{journal}{British Journal of Disorders of Communication} \bibinfo{volume}{15} (\bibinfo{year}{1980}) \bibinfo{pages}{165--173}.
\bibitem[{Yorkston et~al.(1984)Yorkston, Beukelman, and Traynor}]{yorkston1984assessment}
\bibinfo{author}{K.~M. Yorkston}, \bibinfo{author}{D.~R. Beukelman}, \bibinfo{author}{C.~Traynor}, \bibinfo{title}{Assessment of intelligibility of dysarthric speech}, \bibinfo{publisher}{Pro-ed Austin, TX}, \bibinfo{year}{1984}.
\bibitem[{Robertson(1987)}]{robertson1987dysarthria}
\bibinfo{author}{S.~J. Robertson}, \bibinfo{title}{Dysarthria profile}, \bibinfo{publisher}{Communication Skill Builders}, \bibinfo{year}{1987}.
\bibitem[{Jacobson et~al.(1997)Jacobson, Johnson, Grywalski, Silbergleit, Jacobson, Benninger, and Newman}]{jacobson1997voice}
\bibinfo{author}{B.~H. Jacobson}, \bibinfo{author}{A.~Johnson}, \bibinfo{author}{C.~Grywalski}, \bibinfo{author}{A.~Silbergleit}, \bibinfo{author}{G.~Jacobson}, \bibinfo{author}{M.~S. Benninger}, \bibinfo{author}{C.~W. Newman},
\newblock \bibinfo{title}{The voice handicap index (vhi) development and validation},
\newblock \bibinfo{journal}{American journal of speech-language pathology} \bibinfo{volume}{6} (\bibinfo{year}{1997}) \bibinfo{pages}{66--70}.
\bibitem[{Freed(2018)}]{freed2018motor}
\bibinfo{author}{D.~B. Freed}, \bibinfo{title}{Motor speech disorders: diagnosis and treatment}, \bibinfo{publisher}{Plural Publishing}, \bibinfo{year}{2018}.
\bibitem[{Blaustein and Bar(1983)}]{blaustein1983reliability}
\bibinfo{author}{S.~Blaustein}, \bibinfo{author}{A.~Bar},
\newblock \bibinfo{title}{Reliability of perceptual voice assessment},
\newblock \bibinfo{journal}{Journal of communication disorders} \bibinfo{volume}{16} (\bibinfo{year}{1983}) \bibinfo{pages}{157--161}.
\bibitem[{Al~Yaari and Almaflehi(2014)}]{al2014validation}
\bibinfo{author}{S.~A.~S. Al~Yaari}, \bibinfo{author}{N.~Almaflehi},
\newblock \bibinfo{title}{Validation of communication activities of daily living-(cadl-2) on arab aphasics: Controlled study},
\newblock \bibinfo{journal}{J Study Engl Linguist} \bibinfo{volume}{2} (\bibinfo{year}{2014}) \bibinfo{pages}{34}.
\bibitem[{Altaher et~al.(2019)Altaher, Chu, Razak et~al.}]{altaher2019report}
\bibinfo{author}{A.~M. Altaher}, \bibinfo{author}{S.~Y. Chu}, \bibinfo{author}{R.~A. Razak}, et~al.,
\newblock \bibinfo{title}{A report of assessment tools for individuals with dysarthria},
\newblock \bibinfo{journal}{The Open Public Health Journal} \bibinfo{volume}{12} (\bibinfo{year}{2019}).
\bibitem[{Godino-Llorente and Gomez-Vilda(2004)}]{godino2004automatic}
\bibinfo{author}{J.~I. Godino-Llorente}, \bibinfo{author}{P.~Gomez-Vilda},
\newblock \bibinfo{title}{Automatic detection of voice impairments by means of short-term cepstral parameters and neural network based detectors},
\newblock \bibinfo{journal}{IEEE Transactions on Biomedical Engineering} \bibinfo{volume}{51} (\bibinfo{year}{2004}) \bibinfo{pages}{380--384}.
\bibitem[{Hernandez et~al.(2020)Hernandez, Kim, and Chung}]{hernandez2020prosody}
\bibinfo{author}{A.~Hernandez}, \bibinfo{author}{S.~Kim}, \bibinfo{author}{M.~Chung},
\newblock \bibinfo{title}{Prosody-based measures for automatic severity assessment of dysarthric speech},
\newblock \bibinfo{journal}{Applied Sciences} \bibinfo{volume}{10} (\bibinfo{year}{2020}) \bibinfo{pages}{6999}.
\bibitem[{Gallardo-Antol{\'\i}n and Montero(2021)}]{gallardo2021auditory}
\bibinfo{author}{A.~Gallardo-Antol{\'\i}n}, \bibinfo{author}{J.~M. Montero},
\newblock \bibinfo{title}{An auditory saliency pooling-based lstm model for speech intelligibility classification},
\newblock \bibinfo{journal}{Symmetry} \bibinfo{volume}{13} (\bibinfo{year}{2021}) \bibinfo{pages}{1728}.
\bibitem[{V{\'a}squez-Correa et~al.(2018)V{\'a}squez-Correa, Arias-Vergara, Orozco-Arroyave, and N{\"o}th}]{vasquez2018multitask}
\bibinfo{author}{J.~C. V{\'a}squez-Correa}, \bibinfo{author}{T.~Arias-Vergara}, \bibinfo{author}{J.~R. Orozco-Arroyave}, \bibinfo{author}{E.~N{\"o}th},
\newblock \bibinfo{title}{A multitask learning approach to assess the dysarthria severity in patients with parkinson's disease.},
\newblock in: \bibinfo{booktitle}{INTERSPEECH}, \bibinfo{year}{2018}, pp. \bibinfo{pages}{456--460}.
\bibitem[{Parisi et~al.(2018)Parisi, RaviChandran, and Manaog}]{parisi2018feature}
\bibinfo{author}{L.~Parisi}, \bibinfo{author}{N.~RaviChandran}, \bibinfo{author}{M.~L. Manaog},
\newblock \bibinfo{title}{Feature-driven machine learning to improve early diagnosis of parkinson's disease},
\newblock \bibinfo{journal}{Expert Systems with Applications} \bibinfo{volume}{110} (\bibinfo{year}{2018}) \bibinfo{pages}{182--190}.
\bibitem[{Xu et~al.(2023)Xu, Liss, and Berisha}]{xu2023dysarthria}
\bibinfo{author}{L.~Xu}, \bibinfo{author}{J.~Liss}, \bibinfo{author}{V.~Berisha},
\newblock \bibinfo{title}{Dysarthria detection based on a deep learning model with a clinically-interpretable layer},
\newblock \bibinfo{journal}{JASA Express Letters} \bibinfo{volume}{3} (\bibinfo{year}{2023}) \bibinfo{pages}{015201}.
\bibitem[{Kent et~al.(1999)Kent, Weismer, Kent, Vorperian, and Duffy}]{kent1999acoustic}
\bibinfo{author}{R.~D. Kent}, \bibinfo{author}{G.~Weismer}, \bibinfo{author}{J.~F. Kent}, \bibinfo{author}{H.~K. Vorperian}, \bibinfo{author}{J.~R. Duffy},
\newblock \bibinfo{title}{Acoustic studies of dysarthric speech: Methods, progress, and potential},
\newblock \bibinfo{journal}{Journal of communication disorders} \bibinfo{volume}{32} (\bibinfo{year}{1999}) \bibinfo{pages}{141--186}.
\bibitem[{Sapir et~al.(2010)Sapir, Ramig, Spielman, and Fox}]{sapir2010formant}
\bibinfo{author}{S.~Sapir}, \bibinfo{author}{L.~O. Ramig}, \bibinfo{author}{J.~L. Spielman}, \bibinfo{author}{C.~Fox},
\newblock \bibinfo{title}{Formant centralization ratio: A proposal for a new acoustic measure of dysarthric speech}  (\bibinfo{year}{2010}).
\bibitem[{Kim et~al.(2011)Kim, Hasegawa-Johnson, and Perlman}]{kim2011vowel}
\bibinfo{author}{H.~Kim}, \bibinfo{author}{M.~Hasegawa-Johnson}, \bibinfo{author}{A.~Perlman},
\newblock \bibinfo{title}{Vowel contrast and speech intelligibility in dysarthria},
\newblock \bibinfo{journal}{Folia Phoniatrica et Logopaedica} \bibinfo{volume}{63} (\bibinfo{year}{2011}) \bibinfo{pages}{187--194}.
\bibitem[{Orozco-Arroyave et~al.(2014)Orozco-Arroyave, Arias-Londo{\~n}o, Vargas-Bonilla, Gonzalez-R{\'a}tiva, and N{\"o}th}]{orozco2014new}
\bibinfo{author}{J.~R. Orozco-Arroyave}, \bibinfo{author}{J.~D. Arias-Londo{\~n}o}, \bibinfo{author}{J.~F. Vargas-Bonilla}, \bibinfo{author}{M.~C. Gonzalez-R{\'a}tiva}, \bibinfo{author}{E.~N{\"o}th},
\newblock \bibinfo{title}{New spanish speech corpus database for the analysis of people suffering from parkinson's disease.},
\newblock in: \bibinfo{booktitle}{LREC}, \bibinfo{year}{2014}, pp. \bibinfo{pages}{342--347}.
\bibitem[{Tartarisco et~al.(2021)Tartarisco, Bruschetta, Summa, Ruta, Favetta, Busa, Romano, Castelli, Marino, Cerasa et~al.}]{tartarisco2021artificial}
\bibinfo{author}{G.~Tartarisco}, \bibinfo{author}{R.~Bruschetta}, \bibinfo{author}{S.~Summa}, \bibinfo{author}{L.~Ruta}, \bibinfo{author}{M.~Favetta}, \bibinfo{author}{M.~Busa}, \bibinfo{author}{A.~Romano}, \bibinfo{author}{E.~Castelli}, \bibinfo{author}{F.~Marino}, \bibinfo{author}{A.~Cerasa}, et~al.,
\newblock \bibinfo{title}{Artificial intelligence for dysarthria assessment in children with ataxia: A hierarchical approach},
\newblock \bibinfo{journal}{IEEE Access} \bibinfo{volume}{9} (\bibinfo{year}{2021}) \bibinfo{pages}{166720--166735}.
\bibitem[{Kempler and Van~Lancker(2002)}]{kempler2002effect}
\bibinfo{author}{D.~Kempler}, \bibinfo{author}{D.~Van~Lancker},
\newblock \bibinfo{title}{Effect of speech task on intelligibility in dysarthria: A case study of parkinson's disease},
\newblock \bibinfo{journal}{Brain and language} \bibinfo{volume}{80} (\bibinfo{year}{2002}) \bibinfo{pages}{449--464}.
\bibitem[{Ayvaz et~al.(2022)Ayvaz, G{\"u}r{\"u}ler, Khan, Ahmed, Whangbo, and Bobomirzaevich}]{ayvaz2022automatic}
\bibinfo{author}{U.~Ayvaz}, \bibinfo{author}{H.~G{\"u}r{\"u}ler}, \bibinfo{author}{F.~Khan}, \bibinfo{author}{N.~Ahmed}, \bibinfo{author}{T.~Whangbo}, \bibinfo{author}{A.~Bobomirzaevich},
\newblock \bibinfo{title}{Automatic speaker recognition using mel-frequency cepstral coefficients through machine learning},
\newblock \bibinfo{journal}{CMC-Computers Materials \& Continua} \bibinfo{volume}{71} (\bibinfo{year}{2022}).
\bibitem[{Joshy and Rajan(2023)}]{joshy2023dysarthria}
\bibinfo{author}{A.~A. Joshy}, \bibinfo{author}{R.~Rajan},
\newblock \bibinfo{title}{Dysarthria severity classification using multi-head attention and multi-task learning},
\newblock \bibinfo{journal}{Speech Communication} \bibinfo{volume}{147} (\bibinfo{year}{2023}) \bibinfo{pages}{1--11}.
\bibitem[{Bandini et~al.(2018)Bandini, Green, Taati, Orlandi, Zinman, and Yunusova}]{bandini2018automatic}
\bibinfo{author}{A.~Bandini}, \bibinfo{author}{J.~R. Green}, \bibinfo{author}{B.~Taati}, \bibinfo{author}{S.~Orlandi}, \bibinfo{author}{L.~Zinman}, \bibinfo{author}{Y.~Yunusova},
\newblock \bibinfo{title}{Automatic detection of amyotrophic lateral sclerosis (als) from video-based analysis of facial movements: speech and non-speech tasks},
\newblock in: \bibinfo{booktitle}{2018 13th IEEE International Conference on Automatic Face \& Gesture Recognition (FG 2018)}, \bibinfo{organization}{IEEE}, \bibinfo{year}{2018}, pp. \bibinfo{pages}{150--157}.
\bibitem[{Ackermann et~al.(1995)Ackermann, Hertrich, and Scharf}]{ackermann1995kinematic}
\bibinfo{author}{H.~Ackermann}, \bibinfo{author}{I.~Hertrich}, \bibinfo{author}{G.~Scharf},
\newblock \bibinfo{title}{Kinematic analysis of lower lip movements in ataxic dysarthria},
\newblock \bibinfo{journal}{Journal of Speech, Language, and Hearing Research} \bibinfo{volume}{38} (\bibinfo{year}{1995}) \bibinfo{pages}{1252--1259}.
\bibitem[{Xue et~al.(2023{\natexlab{a}})Xue, van Hout, Cucchiarini, and Strik}]{xue2023assessing}
\bibinfo{author}{W.~Xue}, \bibinfo{author}{R.~van Hout}, \bibinfo{author}{C.~Cucchiarini}, \bibinfo{author}{H.~Strik},
\newblock \bibinfo{title}{Assessing speech intelligibility of pathological speech in sentences and word lists: The contribution of phoneme-level measures},
\newblock \bibinfo{journal}{Journal of Communication Disorders} \bibinfo{volume}{102} (\bibinfo{year}{2023}{\natexlab{a}}) \bibinfo{pages}{106301}.
\bibitem[{Xue et~al.(2023{\natexlab{b}})Xue, van Hout, Cucchiarini, and Strik}]{xue2023assessing2}
\bibinfo{author}{W.~Xue}, \bibinfo{author}{R.~van Hout}, \bibinfo{author}{C.~Cucchiarini}, \bibinfo{author}{H.~Strik},
\newblock \bibinfo{title}{Assessing speech intelligibility of pathological speech: test types, ratings and transcription measures},
\newblock \bibinfo{journal}{Clinical Linguistics \& Phonetics} \bibinfo{volume}{37} (\bibinfo{year}{2023}{\natexlab{b}}) \bibinfo{pages}{52--76}.
\bibitem[{Lehner and Ziegler(2021)}]{lehner2021impact}
\bibinfo{author}{K.~Lehner}, \bibinfo{author}{W.~Ziegler},
\newblock \bibinfo{title}{The impact of lexical and articulatory factors in the automatic selection of test materials for a web-based assessment of intelligibility in dysarthria},
\newblock \bibinfo{journal}{Journal of Speech, Language, and Hearing Research} \bibinfo{volume}{64} (\bibinfo{year}{2021}) \bibinfo{pages}{2196--2212}.
\bibitem[{Japkowicz and Shah(2011)}]{japkowicz2011evaluating}
\bibinfo{author}{N.~Japkowicz}, \bibinfo{author}{M.~Shah}, \bibinfo{title}{Evaluating learning algorithms: a classification perspective}, \bibinfo{publisher}{Cambridge University Press}, \bibinfo{year}{2011}.
\bibitem[{Powers(2020)}]{powers2020evaluation}
\bibinfo{author}{D.~M. Powers},
\newblock \bibinfo{title}{Evaluation: from precision, recall and f-measure to roc, informedness, markedness and correlation},
\newblock \bibinfo{journal}{arXiv preprint arXiv:2010.16061}  (\bibinfo{year}{2020}).
\bibitem[{Bradley(1997)}]{bradley1997use}
\bibinfo{author}{A.~P. Bradley},
\newblock \bibinfo{title}{The use of the area under the roc curve in the evaluation of machine learning algorithms},
\newblock \bibinfo{journal}{Pattern recognition} \bibinfo{volume}{30} (\bibinfo{year}{1997}) \bibinfo{pages}{1145--1159}.
\bibitem[{Hastie et~al.(2009)Hastie, Tibshirani, Friedman, and Friedman}]{hastie2009elements}
\bibinfo{author}{T.~Hastie}, \bibinfo{author}{R.~Tibshirani}, \bibinfo{author}{J.~H. Friedman}, \bibinfo{author}{J.~H. Friedman}, \bibinfo{title}{The elements of statistical learning: data mining, inference, and prediction}, volume~\bibinfo{volume}{2}, \bibinfo{publisher}{Springer}, \bibinfo{year}{2009}.
\bibitem[{Rudzicz et~al.(2012)Rudzicz, Namasivayam, and Wolff}]{rudzicz2012torgo}
\bibinfo{author}{F.~Rudzicz}, \bibinfo{author}{A.~K. Namasivayam}, \bibinfo{author}{T.~Wolff},
\newblock \bibinfo{title}{The torgo database of acoustic and articulatory speech from speakers with dysarthria},
\newblock \bibinfo{journal}{Language Resources and Evaluation} \bibinfo{volume}{46} (\bibinfo{year}{2012}) \bibinfo{pages}{523--541}.
\bibitem[{Kim et~al.(2008)Kim, Hasegawa-Johnson, Perlman, Gunderson, Huang, Watkin, and Frame}]{kim2008dysarthric}
\bibinfo{author}{H.~Kim}, \bibinfo{author}{M.~Hasegawa-Johnson}, \bibinfo{author}{A.~Perlman}, \bibinfo{author}{J.~Gunderson}, \bibinfo{author}{T.~S. Huang}, \bibinfo{author}{K.~Watkin}, \bibinfo{author}{S.~Frame},
\newblock \bibinfo{title}{Dysarthric speech database for universal access research},
\newblock in: \bibinfo{booktitle}{Ninth Annual Conference of the International Speech Communication Association}, \bibinfo{year}{2008}.
\bibitem[{Choi et~al.(2012)Choi, Kim, Kim, Lee, Um, and Chung}]{choi2012dysarthric}
\bibinfo{author}{D.-L. Choi}, \bibinfo{author}{B.-W. Kim}, \bibinfo{author}{Y.-W. Kim}, \bibinfo{author}{Y.-J. Lee}, \bibinfo{author}{Y.~Um}, \bibinfo{author}{M.~Chung},
\newblock \bibinfo{title}{Dysarthric speech database for development of qolt software technology.},
\newblock in: \bibinfo{booktitle}{LREC}, \bibinfo{year}{2012}, pp. \bibinfo{pages}{3378--3381}.
\bibitem[{Menendez-Pidal et~al.(1996)Menendez-Pidal, Polikoff, Peters, Leonzio, and Bunnell}]{menendez1996nemours}
\bibinfo{author}{X.~Menendez-Pidal}, \bibinfo{author}{J.~B. Polikoff}, \bibinfo{author}{S.~M. Peters}, \bibinfo{author}{J.~E. Leonzio}, \bibinfo{author}{H.~T. Bunnell},
\newblock \bibinfo{title}{The nemours database of dysarthric speech},
\newblock in: \bibinfo{booktitle}{Proceeding of Fourth International Conference on Spoken Language Processing. ICSLP'96}, volume~\bibinfo{volume}{3}, \bibinfo{organization}{IEEE}, \bibinfo{year}{1996}, pp. \bibinfo{pages}{1962--1965}.
\bibitem[{Kent and Hustad(2009)}]{kent2009speech}
\bibinfo{author}{R.~Kent}, \bibinfo{author}{K.~Hustad},
\newblock \bibinfo{title}{Speech production: development}  (\bibinfo{year}{2009}).
\bibitem[{L{\'o}pez-de Ipi{\~n}a et~al.(2013)L{\'o}pez-de Ipi{\~n}a, Alonso, Travieso, Sol{\'e}-Casals, Egiraun, Faundez-Zanuy, Ezeiza, Barroso, Ecay-Torres, Martinez-Lage et~al.}]{lopez2013selection}
\bibinfo{author}{K.~L{\'o}pez-de Ipi{\~n}a}, \bibinfo{author}{J.-B. Alonso}, \bibinfo{author}{C.~M. Travieso}, \bibinfo{author}{J.~Sol{\'e}-Casals}, \bibinfo{author}{H.~Egiraun}, \bibinfo{author}{M.~Faundez-Zanuy}, \bibinfo{author}{A.~Ezeiza}, \bibinfo{author}{N.~Barroso}, \bibinfo{author}{M.~Ecay-Torres}, \bibinfo{author}{P.~Martinez-Lage}, et~al.,
\newblock \bibinfo{title}{On the selection of non-invasive methods based on speech analysis oriented to automatic alzheimer disease diagnosis},
\newblock \bibinfo{journal}{Sensors} \bibinfo{volume}{13} (\bibinfo{year}{2013}) \bibinfo{pages}{6730--6745}.
\bibitem[{Luz(2017)}]{luz2017longitudinal}
\bibinfo{author}{S.~Luz},
\newblock \bibinfo{title}{Longitudinal monitoring and detection of alzheimer's type dementia from spontaneous speech data},
\newblock in: \bibinfo{booktitle}{2017 IEEE 30th International Symposium on Computer-Based Medical Systems (CBMS)}, \bibinfo{organization}{IEEE}, \bibinfo{year}{2017}, pp. \bibinfo{pages}{45--46}.
\bibitem[{Turner et~al.(1995)Turner, Tjaden, and Weismer}]{turner1995influence}
\bibinfo{author}{G.~S. Turner}, \bibinfo{author}{K.~Tjaden}, \bibinfo{author}{G.~Weismer},
\newblock \bibinfo{title}{The influence of speaking rate on vowel space and speech intelligibility for individuals with amyotrophic lateral sclerosis},
\newblock \bibinfo{journal}{Journal of Speech, Language, and Hearing Research} \bibinfo{volume}{38} (\bibinfo{year}{1995}) \bibinfo{pages}{1001--1013}.
\bibitem[{Weismer et~al.(2001)Weismer, Jeng, Laures, Kent, and Kent}]{weismer2001acoustic}
\bibinfo{author}{G.~Weismer}, \bibinfo{author}{J.-Y. Jeng}, \bibinfo{author}{J.~S. Laures}, \bibinfo{author}{R.~D. Kent}, \bibinfo{author}{J.~F. Kent},
\newblock \bibinfo{title}{Acoustic and intelligibility characteristics of sentence production in neurogenic speech disorders},
\newblock \bibinfo{journal}{Folia Phoniatrica et Logopaedica} \bibinfo{volume}{53} (\bibinfo{year}{2001}) \bibinfo{pages}{1--18}.
\bibitem[{Modale et~al.(2020)Modale, Sable, and Deshmukh}]{modale2020review}
\bibinfo{author}{S.~Modale}, \bibinfo{author}{G.~Sable}, \bibinfo{author}{R.~Deshmukh},
\newblock \bibinfo{title}{A review: Devnagri speech to text for marathwada region},
\newblock in: \bibinfo{booktitle}{Proceedings of International Conference on Wireless Communication: ICWiCOM 2019}, \bibinfo{organization}{Springer}, \bibinfo{year}{2020}, pp. \bibinfo{pages}{525--532}.
\bibitem[{Joshy and Rajan(2022)}]{joshy2022automated}
\bibinfo{author}{A.~A. Joshy}, \bibinfo{author}{R.~Rajan},
\newblock \bibinfo{title}{Automated dysarthria severity classification: A study on acoustic features and deep learning techniques},
\newblock \bibinfo{journal}{IEEE Transactions on Neural Systems and Rehabilitation Engineering} \bibinfo{volume}{30} (\bibinfo{year}{2022}) \bibinfo{pages}{1147--1157}.
\bibitem[{Yeo et~al.(2023)Yeo, Choi, Kim, and Chung}]{yeo2023automatic}
\bibinfo{author}{E.~J. Yeo}, \bibinfo{author}{K.~Choi}, \bibinfo{author}{S.~Kim}, \bibinfo{author}{M.~Chung},
\newblock \bibinfo{title}{Automatic severity classification of dysarthric speech by using self-supervised model with multi-task learning},
\newblock in: \bibinfo{booktitle}{ICASSP 2023-2023 IEEE International Conference on Acoustics, Speech and Signal Processing (ICASSP)}, \bibinfo{organization}{IEEE}, \bibinfo{year}{2023}, pp. \bibinfo{pages}{1--5}.
\bibitem[{Karjigi et~al.(2023)Karjigi, Sreedevi et~al.}]{karjigi2023speech}
\bibinfo{author}{V.~Karjigi}, \bibinfo{author}{N.~Sreedevi}, et~al.,
\newblock \bibinfo{title}{Speech intelligibility assessment of dysarthria using fisher vector encoding},
\newblock \bibinfo{journal}{Computer Speech \& Language} \bibinfo{volume}{77} (\bibinfo{year}{2023}) \bibinfo{pages}{101411}.
\bibitem[{Yeo et~al.(2021)Yeo, Kim, and Chung}]{yeo2021automatic}
\bibinfo{author}{E.~J. Yeo}, \bibinfo{author}{S.~Kim}, \bibinfo{author}{M.~Chung},
\newblock \bibinfo{title}{Automatic severity classification of korean dysarthric speech using phoneme-level pronunciation features.},
\newblock in: \bibinfo{booktitle}{Interspeech}, \bibinfo{year}{2021}, pp. \bibinfo{pages}{4838--4842}.
\bibitem[{Kadi et~al.(2014)Kadi, Selouani, Boudraa, and Boudraa}]{kadi2014automated}
\bibinfo{author}{K.~L. Kadi}, \bibinfo{author}{S.~A. Selouani}, \bibinfo{author}{B.~Boudraa}, \bibinfo{author}{M.~Boudraa},
\newblock \bibinfo{title}{Automated diagnosis and assessment of dysarthric speech using relevant prosodic features},
\newblock in: \bibinfo{booktitle}{Transactions on Engineering Technologies: Special Volume of the World Congress on Engineering 2013}, \bibinfo{organization}{Springer}, \bibinfo{year}{2014}, pp. \bibinfo{pages}{529--542}.
\bibitem[{Al-Qatab and Mustafa(2021)}]{al2021classification}
\bibinfo{author}{B.~A. Al-Qatab}, \bibinfo{author}{M.~B. Mustafa},
\newblock \bibinfo{title}{Classification of dysarthric speech according to the severity of impairment: an analysis of acoustic features},
\newblock \bibinfo{journal}{IEEE Access} \bibinfo{volume}{9} (\bibinfo{year}{2021}) \bibinfo{pages}{18183--18194}.
\bibitem[{Bhat et~al.(2017)Bhat, Vachhani, and Kopparapu}]{bhat2017automatic}
\bibinfo{author}{C.~Bhat}, \bibinfo{author}{B.~Vachhani}, \bibinfo{author}{S.~K. Kopparapu},
\newblock \bibinfo{title}{Automatic assessment of dysarthria severity level using audio descriptors},
\newblock in: \bibinfo{booktitle}{2017 IEEE International Conference on Acoustics, Speech and Signal Processing (ICASSP)}, \bibinfo{organization}{IEEE}, \bibinfo{year}{2017}, pp. \bibinfo{pages}{5070--5074}.
\bibitem[{Hernandez et~al.(2020)Hernandez, Yeo, Kim, and Chung}]{hernandez2020dysarthria}
\bibinfo{author}{A.~Hernandez}, \bibinfo{author}{E.~J. Yeo}, \bibinfo{author}{S.~Kim}, \bibinfo{author}{M.~Chung},
\newblock \bibinfo{title}{Dysarthria detection and severity assessment using rhythm-based metrics.},
\newblock in: \bibinfo{booktitle}{INTERSPEECH}, \bibinfo{year}{2020}, pp. \bibinfo{pages}{2897--2901}.
\bibitem[{Kachhi et~al.(2022)Kachhi, Therattil, Patil, Sailor, and Patil}]{kachhi2022teager}
\bibinfo{author}{A.~Kachhi}, \bibinfo{author}{A.~Therattil}, \bibinfo{author}{A.~T. Patil}, \bibinfo{author}{H.~B. Sailor}, \bibinfo{author}{H.~A. Patil},
\newblock \bibinfo{title}{Teager energy cepstral coefficients for classification of dysarthric speech severity-level},
\newblock in: \bibinfo{booktitle}{2022 Asia-Pacific Signal and Information Processing Association Annual Summit and Conference (APSIPA ASC)}, \bibinfo{organization}{IEEE}, \bibinfo{year}{2022}, pp. \bibinfo{pages}{1462--1468}.
\bibitem[{Paja and Falk(2012)}]{paja2012automated}
\bibinfo{author}{M.~S. Paja}, \bibinfo{author}{T.~H. Falk},
\newblock \bibinfo{title}{Automated dysarthria severity classification for improved objective intelligibility assessment of spastic dysarthric speech},
\newblock in: \bibinfo{booktitle}{Thirteenth Annual Conference of the International Speech Communication Association}, \bibinfo{year}{2012}.
\bibitem[{Yeo et~al.(2022)Yeo, Choi, Kim, and Chung}]{yeo2022cross}
\bibinfo{author}{E.~J. Yeo}, \bibinfo{author}{K.~Choi}, \bibinfo{author}{S.~Kim}, \bibinfo{author}{M.~Chung},
\newblock \bibinfo{title}{Cross-lingual dysarthria severity classification for english, korean, and tamil},
\newblock in: \bibinfo{booktitle}{2022 Asia-Pacific Signal and Information Processing Association Annual Summit and Conference (APSIPA ASC)}, \bibinfo{organization}{IEEE}, \bibinfo{year}{2022}, pp. \bibinfo{pages}{566--574}.
\bibitem[{Kadi et~al.(2013)Kadi, Selouani, Boudraa, and Boudraa}]{kadi2013discriminative}
\bibinfo{author}{K.~Kadi}, \bibinfo{author}{S.~Selouani}, \bibinfo{author}{B.~Boudraa}, \bibinfo{author}{M.~Boudraa},
\newblock \bibinfo{title}{Discriminative prosodic features to assess the dysarthria severity levels},
\newblock in: \bibinfo{booktitle}{Proceedings of the World Congress on Engineering}, volume~\bibinfo{volume}{3}, \bibinfo{year}{2013}.
\bibitem[{Vyas et~al.(2016)Vyas, Dutta, Prinosil, and Har{\'a}r}]{vyas2016automatic}
\bibinfo{author}{G.~Vyas}, \bibinfo{author}{M.~K. Dutta}, \bibinfo{author}{J.~Prinosil}, \bibinfo{author}{P.~Har{\'a}r},
\newblock \bibinfo{title}{An automatic diagnosis and assessment of dysarthric speech using speech disorder specific prosodic features},
\newblock in: \bibinfo{booktitle}{2016 39th International Conference on Telecommunications and Signal Processing (TSP)}, \bibinfo{organization}{IEEE}, \bibinfo{year}{2016}, pp. \bibinfo{pages}{515--518}.
\bibitem[{Kim et~al.(2013)Kim, Yoo, and Kim}]{kim2013dysarthric}
\bibinfo{author}{M.~J. Kim}, \bibinfo{author}{J.~Yoo}, \bibinfo{author}{H.~Kim},
\newblock \bibinfo{title}{Dysarthric speech recognition using dysarthria-severity-dependent and speaker-adaptive models.},
\newblock in: \bibinfo{booktitle}{Interspeech}, \bibinfo{year}{2013}, pp. \bibinfo{pages}{3622--3626}.
\bibitem[{Purohit et~al.(2021)Purohit, Parmar, Patel, Malaviya, and Patii}]{purohit2021weak}
\bibinfo{author}{M.~Purohit}, \bibinfo{author}{M.~Parmar}, \bibinfo{author}{M.~Patel}, \bibinfo{author}{H.~Malaviya}, \bibinfo{author}{H.~A. Patii},
\newblock \bibinfo{title}{Weak speech supervision: A case study of dysarthria severity classification},
\newblock in: \bibinfo{booktitle}{2020 28th European Signal Processing Conference (EUSIPCO)}, \bibinfo{organization}{IEEE}, \bibinfo{year}{2021}, pp. \bibinfo{pages}{101--105}.
\bibitem[{Kachhi et~al.(2022)Kachhi, Therattil, Patil, Sailor, and Patil}]{kachhi2022significance}
\bibinfo{author}{A.~Kachhi}, \bibinfo{author}{A.~Therattil}, \bibinfo{author}{A.~T. Patil}, \bibinfo{author}{H.~B. Sailor}, \bibinfo{author}{H.~A. Patil},
\newblock \bibinfo{title}{Significance of energy features for severity classification of dysarthria},
\newblock in: \bibinfo{booktitle}{Speech and Computer: 24th International Conference, SPECOM 2022, Gurugram, India, November 14--16, 2022, Proceedings}, \bibinfo{organization}{Springer}, \bibinfo{year}{2022}, pp. \bibinfo{pages}{325--337}.
\bibitem[{Mendoza~Ramos et~al.(2021)Mendoza~Ramos, Lowit, Van~den Steen, Kairuz Hernandez-Diaz, Hernandez-Diaz~Huici, De~Bodt, and Van~Nuffelen}]{mendoza2021acoustic}
\bibinfo{author}{V.~Mendoza~Ramos}, \bibinfo{author}{A.~Lowit}, \bibinfo{author}{L.~Van~den Steen}, \bibinfo{author}{H.~A. Kairuz Hernandez-Diaz}, \bibinfo{author}{M.~E. Hernandez-Diaz~Huici}, \bibinfo{author}{M.~De~Bodt}, \bibinfo{author}{G.~Van~Nuffelen},
\newblock \bibinfo{title}{Acoustic identification of sentence accent in speakers with dysarthria: cross-population validation and severity related patterns},
\newblock \bibinfo{journal}{Brain Sciences} \bibinfo{volume}{11} (\bibinfo{year}{2021}) \bibinfo{pages}{1344}.
\bibitem[{Gurugubelli and Vuppala(2020)}]{gurugubelli2020analytic}
\bibinfo{author}{K.~Gurugubelli}, \bibinfo{author}{A.~K. Vuppala},
\newblock \bibinfo{title}{Analytic phase features for dysarthric speech detection and intelligibility assessment},
\newblock \bibinfo{journal}{Speech Communication} \bibinfo{volume}{121} (\bibinfo{year}{2020}) \bibinfo{pages}{1--15}.
\bibitem[{Joshy and Rajan(2021)}]{joshy2021automated}
\bibinfo{author}{A.~A. Joshy}, \bibinfo{author}{R.~Rajan},
\newblock \bibinfo{title}{Automated dysarthria severity classification using deep learning frameworks},
\newblock in: \bibinfo{booktitle}{2020 28th European Signal Processing Conference (EUSIPCO)}, \bibinfo{organization}{IEEE}, \bibinfo{year}{2021}, pp. \bibinfo{pages}{116--120}.
\bibitem[{Narendra and Alku(2020)}]{narendra2020automatic}
\bibinfo{author}{N.~P. Narendra}, \bibinfo{author}{P.~Alku},
\newblock \bibinfo{title}{Automatic intelligibility assessment of dysarthric speech using glottal parameters},
\newblock \bibinfo{journal}{Speech Communication} \bibinfo{volume}{123} (\bibinfo{year}{2020}) \bibinfo{pages}{1--9}.
\bibitem[{Gillespie et~al.(2017)Gillespie, Logan, Moore, Laures-Gore, Russell, and Patel}]{gillespie2017cross}
\bibinfo{author}{S.~Gillespie}, \bibinfo{author}{Y.-Y. Logan}, \bibinfo{author}{E.~Moore}, \bibinfo{author}{J.~Laures-Gore}, \bibinfo{author}{S.~Russell}, \bibinfo{author}{R.~Patel},
\newblock \bibinfo{title}{Cross-database models for the classification of dysarthria presence.},
\newblock in: \bibinfo{booktitle}{Interspeech}, \bibinfo{year}{2017}, pp. \bibinfo{pages}{3127--3131}.
\bibitem[{Kadi et~al.(2016)Kadi, Selouani, Boudraa, and Boudraa}]{kadi2016fully}
\bibinfo{author}{K.~L. Kadi}, \bibinfo{author}{S.~A. Selouani}, \bibinfo{author}{B.~Boudraa}, \bibinfo{author}{M.~Boudraa},
\newblock \bibinfo{title}{Fully automated speaker identification and intelligibility assessment in dysarthria disease using auditory knowledge},
\newblock \bibinfo{journal}{Biocybernetics and Biomedical Engineering} \bibinfo{volume}{36} (\bibinfo{year}{2016}) \bibinfo{pages}{233--247}.
\bibitem[{Dahmani et~al.(2014)Dahmani, Selouani, Doghmane, O’Shaughnessy, and Chetouani}]{dahmani2014relevance}
\bibinfo{author}{H.~Dahmani}, \bibinfo{author}{S.-A. Selouani}, \bibinfo{author}{N.~Doghmane}, \bibinfo{author}{D.~O’Shaughnessy}, \bibinfo{author}{M.~Chetouani},
\newblock \bibinfo{title}{On the relevance of using rhythmic metrics and svm to assess dysarthric severity},
\newblock \bibinfo{journal}{International Journal of Biometrics} \bibinfo{volume}{6} (\bibinfo{year}{2014}) \bibinfo{pages}{248--271}.
\bibitem[{Joshy et~al.(2023)Joshy, Parameswaran, Nair, and Rajan}]{joshy2023statistical}
\bibinfo{author}{A.~A. Joshy}, \bibinfo{author}{P.~Parameswaran}, \bibinfo{author}{S.~R. Nair}, \bibinfo{author}{R.~Rajan},
\newblock \bibinfo{title}{Statistical analysis of speech disorder specific features to characterise dysarthria severity level},
\newblock in: \bibinfo{booktitle}{ICASSP 2023-2023 IEEE International Conference on Acoustics, Speech and Signal Processing (ICASSP)}, \bibinfo{organization}{IEEE}, \bibinfo{year}{2023}, pp. \bibinfo{pages}{1--5}.
\bibitem[{Javanmardi et~al.(2023)Javanmardi, Tirronen, Kodali, Kadiri, and Alku}]{javanmardi2023wav2vec}
\bibinfo{author}{F.~Javanmardi}, \bibinfo{author}{S.~Tirronen}, \bibinfo{author}{M.~Kodali}, \bibinfo{author}{S.~R. Kadiri}, \bibinfo{author}{P.~Alku},
\newblock \bibinfo{title}{Wav2vec-based detection and severity level classification of dysarthria from speech},
\newblock in: \bibinfo{booktitle}{ICASSP 2023-2023 IEEE International Conference on Acoustics, Speech and Signal Processing (ICASSP)}, \bibinfo{organization}{IEEE}, \bibinfo{year}{2023}, pp. \bibinfo{pages}{1--5}.
\bibitem[{DeMino et~al.(2011)DeMino, Kubichek, and Caves}]{demino2011assessing}
\bibinfo{author}{A.~DeMino}, \bibinfo{author}{R.~Kubichek}, \bibinfo{author}{K.~Caves},
\newblock \bibinfo{title}{Assessing dysarthria severity using global statistics and boosting},
\newblock in: \bibinfo{booktitle}{2011 Conference Record of the Forty Fifth Asilomar Conference on Signals, Systems and Computers (ASILOMAR)}, \bibinfo{organization}{IEEE}, \bibinfo{year}{2011}, pp. \bibinfo{pages}{1103--1106}.
\bibitem[{Kachhi et~al.(2022)Kachhi, Therattil, Gupta, and Patil}]{kachhi2022continuous}
\bibinfo{author}{A.~Kachhi}, \bibinfo{author}{A.~Therattil}, \bibinfo{author}{P.~Gupta}, \bibinfo{author}{H.~A. Patil},
\newblock \bibinfo{title}{Continuous wavelet transform for severity-level classification of dysarthria},
\newblock in: \bibinfo{booktitle}{Speech and Computer: 24th International Conference, SPECOM 2022, Gurugram, India, November 14--16, 2022, Proceedings}, \bibinfo{organization}{Springer}, \bibinfo{year}{2022}, pp. \bibinfo{pages}{312--324}.
\bibitem[{Kumar et~al.(2022)Kumar, Keerthivasan, Sasikala et~al.}]{kumar2022towards}
\bibinfo{author}{S.~A. Kumar}, \bibinfo{author}{T.~Keerthivasan}, \bibinfo{author}{S.~Sasikala}, et~al.,
\newblock \bibinfo{title}{Towards improving the performance of dysarthric speech severity assessment system},
\newblock in: \bibinfo{booktitle}{2022 International Conference on Computer Communication and Informatics (ICCCI)}, \bibinfo{organization}{IEEE}, \bibinfo{year}{2022}, pp. \bibinfo{pages}{1--6}.
\bibitem[{Lee et~al.(2019)Lee, Kim, Seo, Oh, Lee, and Leigh}]{lee2019assessment}
\bibinfo{author}{S.~H. Lee}, \bibinfo{author}{M.~Kim}, \bibinfo{author}{H.~G. Seo}, \bibinfo{author}{B.-M. Oh}, \bibinfo{author}{G.~Lee}, \bibinfo{author}{J.-H. Leigh},
\newblock \bibinfo{title}{Assessment of dysarthria using one-word speech recognition with hidden markov models},
\newblock \bibinfo{journal}{Journal of Korean medical science} \bibinfo{volume}{34} (\bibinfo{year}{2019}).
\bibitem[{Joshy and Rajan(2023)}]{joshy2023dysarthria2}
\bibinfo{author}{A.~A. Joshy}, \bibinfo{author}{R.~Rajan},
\newblock \bibinfo{title}{Dysarthria severity assessment using squeeze-and-excitation networks},
\newblock \bibinfo{journal}{Biomedical Signal Processing and Control} \bibinfo{volume}{82} (\bibinfo{year}{2023}) \bibinfo{pages}{104606}.
\bibitem[{Chandrashekar et~al.(2020)Chandrashekar, Karjigi, and Sreedevi}]{chandrashekar2020investigation}
\bibinfo{author}{H.~Chandrashekar}, \bibinfo{author}{V.~Karjigi}, \bibinfo{author}{N.~Sreedevi},
\newblock \bibinfo{title}{Investigation of different time-frequency representations for intelligibility assessment of dysarthric speech},
\newblock \bibinfo{journal}{Ieee transactions on neural systems and rehabilitation engineering} \bibinfo{volume}{28} (\bibinfo{year}{2020}) \bibinfo{pages}{2880--2889}.
\bibitem[{Suhas et~al.(2020)Suhas, Mallela, Illa, Yamini, Atchayaram, Yadav, Gope, and Ghosh}]{suhas2020speech}
\bibinfo{author}{B.~Suhas}, \bibinfo{author}{J.~Mallela}, \bibinfo{author}{A.~Illa}, \bibinfo{author}{B.~Yamini}, \bibinfo{author}{N.~Atchayaram}, \bibinfo{author}{R.~Yadav}, \bibinfo{author}{D.~Gope}, \bibinfo{author}{P.~K. Ghosh},
\newblock \bibinfo{title}{Speech task based automatic classification of als and parkinson’s disease and their severity using log mel spectrograms},
\newblock in: \bibinfo{booktitle}{2020 international conference on signal processing and communications (SPCOM)}, \bibinfo{organization}{IEEE}, \bibinfo{year}{2020}, pp. \bibinfo{pages}{1--5}.
\bibitem[{Fern{\'a}ndez-D{\'\i}az and Gallardo-Antol{\'\i}n(2020)}]{fernandez2020attention}
\bibinfo{author}{M.~Fern{\'a}ndez-D{\'\i}az}, \bibinfo{author}{A.~Gallardo-Antol{\'\i}n},
\newblock \bibinfo{title}{An attention long short-term memory based system for automatic classification of speech intelligibility},
\newblock \bibinfo{journal}{Engineering Applications of Artificial Intelligence} \bibinfo{volume}{96} (\bibinfo{year}{2020}) \bibinfo{pages}{103976}.
\bibitem[{Montalbo(2023)}]{montalbo4442941dysarnet}
\bibinfo{author}{F.~J. Montalbo},
\newblock \bibinfo{title}{Dysarnet: A densely squeezed-and-excited attention-gated residual deep learning model for dysarthric speech recognition and severity estimation},
\newblock \bibinfo{journal}{Available at SSRN 4442941}  (\bibinfo{year}{January 2023}).
\bibitem[{Gupta et~al.(2021)Gupta, Patil, Purohit, Parmar, Patel, Patil, and Guido}]{gupta2021residual}
\bibinfo{author}{S.~Gupta}, \bibinfo{author}{A.~T. Patil}, \bibinfo{author}{M.~Purohit}, \bibinfo{author}{M.~Parmar}, \bibinfo{author}{M.~Patel}, \bibinfo{author}{H.~A. Patil}, \bibinfo{author}{R.~C. Guido},
\newblock \bibinfo{title}{Residual neural network precisely quantifies dysarthria severity-level based on short-duration speech segments},
\newblock \bibinfo{journal}{Neural Networks} \bibinfo{volume}{139} (\bibinfo{year}{2021}) \bibinfo{pages}{105--117}.
\bibitem[{Chandrashekar et~al.(2019)Chandrashekar, Karjigi, and Sreedevi}]{chandrashekar2019spectro}
\bibinfo{author}{H.~Chandrashekar}, \bibinfo{author}{V.~Karjigi}, \bibinfo{author}{N.~Sreedevi},
\newblock \bibinfo{title}{Spectro-temporal representation of speech for intelligibility assessment of dysarthria},
\newblock \bibinfo{journal}{IEEE Journal of Selected Topics in Signal Processing} \bibinfo{volume}{14} (\bibinfo{year}{2019}) \bibinfo{pages}{390--399}.
\bibitem[{Akko{\c{c}}(2012)}]{akkocc2012empirical}
\bibinfo{author}{S.~Akko{\c{c}}},
\newblock \bibinfo{title}{An empirical comparison of conventional techniques, neural networks and the three stage hybrid adaptive neuro fuzzy inference system (anfis) model for credit scoring analysis: The case of turkish credit card data},
\newblock \bibinfo{journal}{European Journal of Operational Research} \bibinfo{volume}{222} (\bibinfo{year}{2012}) \bibinfo{pages}{168--178}.
\bibitem[{Subasi(2007)}]{subasi2007application}
\bibinfo{author}{A.~Subasi},
\newblock \bibinfo{title}{Application of adaptive neuro-fuzzy inference system for epileptic seizure detection using wavelet feature extraction},
\newblock \bibinfo{journal}{Computers in biology and medicine} \bibinfo{volume}{37} (\bibinfo{year}{2007}) \bibinfo{pages}{227--244}.
\bibitem[{Geman(2011)}]{geman2011data}
\bibinfo{author}{O.~Geman},
\newblock \bibinfo{title}{Data processing for parkinson's disease: Tremor, speech and gait signal analysis},
\newblock in: \bibinfo{booktitle}{2011 E-Health and Bioengineering Conference (EHB)}, \bibinfo{organization}{IEEE}, \bibinfo{year}{2011}, pp. \bibinfo{pages}{1--4}.
\bibitem[{Accardo and Mumolo(1998)}]{accardo1998algorithm}
\bibinfo{author}{A.~P. Accardo}, \bibinfo{author}{E.~Mumolo},
\newblock \bibinfo{title}{An algorithm for the automatic differentiation between the speech of normals and patients with friedreich's ataxia based on the short-time fractal dimension},
\newblock \bibinfo{journal}{Computers in biology and medicine} \bibinfo{volume}{28} (\bibinfo{year}{1998}) \bibinfo{pages}{75--89}.
\bibitem[{Rudzicz(2010)}]{rudzicz2010towards}
\bibinfo{author}{F.~Rudzicz},
\newblock \bibinfo{title}{Towards a noisy-channel model of dysarthria in speech recognition},
\newblock in: \bibinfo{booktitle}{Proceedings of the NAACL HLT 2010 Workshop on Speech and Language Processing for Assistive Technologies}, \bibinfo{year}{2010}, pp. \bibinfo{pages}{80--88}.
\bibitem[{Jackson et~al.(2016)Jackson, Tiede, Riley, and Whalen}]{jackson2016recurrence}
\bibinfo{author}{E.~S. Jackson}, \bibinfo{author}{M.~Tiede}, \bibinfo{author}{M.~A. Riley}, \bibinfo{author}{D.~Whalen},
\newblock \bibinfo{title}{Recurrence quantification analysis of sentence-level speech kinematics},
\newblock \bibinfo{journal}{Journal of Speech, Language, and Hearing Research} \bibinfo{volume}{59} (\bibinfo{year}{2016}) \bibinfo{pages}{1315--1326}.
\bibitem[{L{\'e}v{\^e}que et~al.(2022)L{\'e}v{\^e}que, Slis, Lancia, Bruneteau, and Fougeron}]{leveque2022acoustic}
\bibinfo{author}{N.~L{\'e}v{\^e}que}, \bibinfo{author}{A.~Slis}, \bibinfo{author}{L.~Lancia}, \bibinfo{author}{G.~Bruneteau}, \bibinfo{author}{C.~Fougeron},
\newblock \bibinfo{title}{Acoustic change over time in spastic and/or flaccid dysarthria in motor neuron diseases},
\newblock \bibinfo{journal}{Journal of Speech, Language, and Hearing Research} \bibinfo{volume}{65} (\bibinfo{year}{2022}) \bibinfo{pages}{1767--1783}.
\bibitem[{Yue et~al.(2022)Yue, Loweimi, and Cvetkovic}]{yue2022raw}
\bibinfo{author}{Z.~Yue}, \bibinfo{author}{E.~Loweimi}, \bibinfo{author}{Z.~Cvetkovic},
\newblock \bibinfo{title}{Raw source and filter modelling for dysarthric speech recognition},
\newblock in: \bibinfo{booktitle}{ICASSP 2022-2022 IEEE International Conference on Acoustics, Speech and Signal Processing (ICASSP)}, \bibinfo{organization}{IEEE}, \bibinfo{year}{2022}, pp. \bibinfo{pages}{7377--7381}.
\bibitem[{Vainio and Altosaar(1998)}]{vainio1998modeling}
\bibinfo{author}{M.~Vainio}, \bibinfo{author}{T.~Altosaar},
\newblock \bibinfo{title}{Modeling the microprosody of pitch and loudness for speech synthesis with neural networks.},
\newblock in: \bibinfo{booktitle}{ICSLP}, \bibinfo{year}{1998}.
\bibitem[{Lee et~al.(2018)Lee, Bell, and Simmons}]{lee2018articulatory}
\bibinfo{author}{J.~Lee}, \bibinfo{author}{M.~Bell}, \bibinfo{author}{Z.~Simmons},
\newblock \bibinfo{title}{Articulatory kinematic characteristics across the dysarthria severity spectrum in individuals with amyotrophic lateral sclerosis},
\newblock \bibinfo{journal}{American Journal of Speech-Language Pathology} \bibinfo{volume}{27} (\bibinfo{year}{2018}) \bibinfo{pages}{258--269}.

\end{thebibliography}

\end{document}